\documentclass[preprint,12pt]{elsarticle}

\usepackage{graphicx}
\usepackage{pgfplots}
\usepackage{float}
\usepackage{amsmath} 
\usepackage{xcolor}
\usepackage[utf8]{inputenc}
\usepackage[T1]{fontenc}
\pgfplotsset{compat=1.18} 

\usepackage{amssymb}

\usepackage{multirow}
\usepackage{subcaption}
\usepackage{comment}
\usepackage{hyperref}

\begin{document}
\begin{frontmatter}

\title{\$PINN - a Domain Decomposition Method for Bayesian Physics-Informed Neural Networks\tnoteref{label1}}
\author{J\'ulia Vicens Figueres\fnref{label2}}
\author{Juliette Vanderhaeghen\fnref{label3}}
\author{Federica Bragone\fnref{label4}}
\author{Kateryna Morozovska\corref{cor1}\fnref{label4}}
\author{Khemraj Shukla\fnref{label5}}
\cortext[cor1]{KTH Royal Insitute of Technology, Malvinas v\"ag 10, 100-44, Stockholm, Sweden. kmor@kth.se}
\affiliation[label2]{organization={University of Barcelona},
            city={Barcelona},
            country={Spain}}
\affiliation[label3]{organization={UCLouvain},
            city={Louvain-la-Neuve},
            country={Belgium}}
\affiliation[label4]{organization={KTH Royal Institute of Technology},
            city={Stockholm},
            country={Sweden}}

\affiliation[label5]{organization={Brown University},
            city={Providence},
            state={RI},
            country={USA}}
\begin{abstract}
Physics-Informed Neural Networks (PINNs) are a  novel computational approach for solving partial differential equations (PDEs) with noisy and sparse initial and boundary data. Although, \textbf{efficient} quantification of epistemic and aleatoric uncertainties in big multi-scale problems remains challenging. We propose \$PINN a novel method of computing global uncertainty in PDEs 
using a Bayesian framework, by combining local Bayesian Physics-Informed Neural Networks (BPINN) with domain decomposition. The solution continuity across subdomains is obtained by imposing the flux continuity across the interface of neighboring subdomains. To demonstrate the effectiveness of \$PINN, we conduct a series of computational experiments on PDEs in 1D and 2D spatial domains. Although we have adopted conservative PINNs (cPINNs), the method can be seamlessly extended to other domain decomposition techniques. The results infer that the proposed method recovers the global uncertainty by computing the local uncertainty exactly more efficiently as the uncertainty in each subdomain can be computed concurrently. The robustness of \$PINN is verified by adding uncorrelated random noise to the training data up to 15\% and testing for different domain sizes.
\end{abstract}
\end{frontmatter}


\section{Introduction}
During recent years, deep learning techniques have been progressing towards becoming more scalable and robust tools for solving complex problems in science and engineering \cite{DL1, DL2}. The development of methods that combine physical knowledge with data-driven methods that fall under the name of Scientific Machine Learning (SciML) has gained more interest in applied disciplines due to their ability to address a wide range of problems \cite{cuomo}. One of the main methods within this field is Physics-Informed Neural Networks (PINNs) \cite{Raissi} that, by embedding the physics described by Ordinary Differential Equations (ODEs) and Partial Differential Equations (PDEs) into the learning process, can solve and discover equations, or in other words be formulated as forward and inverse problems. The ability of PINNs to combine existing knowledge of the underlying systems using the observed data can potentially improve computational efficiency \cite{cuomo, markidis}.


However, despite their popularity, PINNs have some training limitations, especially when handling noisy data or dealing with multi-scale problems. 
Therefore, two main challenges to be addressed are the uncertainty quantification (UQ) and scalability of PINNs~\cite{PSAROS2023111902}. 
Traditionally, data-driven models experience two types of uncertainties: epistemic and aleatoric. Epistemic uncertainty arises from the model and can be caused by random initialization of parameters, uncertainty in the training process, or lack of data pre-processing. However, there are methods for addressing epistemic uncertainty, such as Deep Ensemble learning \cite{ganaie2022ensemble}, which allows averaging the results over several training iterations to ensure the validity of the outcome. Unlike epistemic uncertainty, the aleatoric uncertainty comes from the stochastic nature of the data itself and cannot be addressed using the same strategies. 

Bayesian inference has been previously used to quantify uncertainties in PDEs \cite{BayesianInference,BayesianInference2}. 
Subsequently, Bayesian PINN (BPINN) \cite{yang2021b} is developed to integrate PINN with Bayesian optimization \cite{Takinghuman} to better handle noisy data. 
The Bayesian Neural Network (BNN) component of the prior adopts Bayesian principles by assigning probability distributions to the weights and biases of the neural network \cite{bishop1997bayesian, bykov2021explaining}. To account for noise in the data, we add noise to the likelihood function. By applying Bayes' rule, we can update the posterior distribution of the model and the parameters of the differential equation. This estimation process enables the propagation of uncertainty from the observed data to the predictions made by the model. BPINN uses iterative sampling to update the posterior distribution driven by the observed data. This iterative process requires multiple iterations to converge to a stable and robust solution, increasing computational costs. Given the marginalization term's intractability for most cases, BPINNs rely on sampling-based algorithms to approximate the posterior. Examples of such methods are Hamiltonian Monte Carlo (HMC) approach~\cite{radivojevic2020modified}, which is an efficient Markov Chain Monte Carlo (MCMC) method~\cite{brooks1998markov, neal2011mcmc, neal2012bayesian} and variational inference~\cite{graves2011practical, blei2017variational}.
 These algorithms generate multiple samples from the posterior distribution, which are used to approximate uncertainty and infer calibrated parameters of neural network and parameters of PDE. 


A second problem that PINNs face and that BPINNs also share is their limitations with larger-scale problems, including complex domains, 3D space domains, the combination of space and time domains, and specific PDEs. As the problem complexity increases, domain decomposition can help PINNs to solve smaller problems. Domain decomposition becomes essential in adapting PINNs to solve multi-scale problems and potentially gaining from parallel programming for solving problems faster \cite{XPINN}. 

Many existing algorithms and methods provide possibilities of domain decomposition for certain types of problems \cite{klawonn}. Generally, domain decomposition methods differ in the strategies for splitting the domain, specifying the interface conditions, and formulating the PINN itself. According to the review in \cite{klawonn}, the split of domain can be divided into 3 types: overlapping domains \cite{FBPINN1, FBPINN2, fBPINN3}, non-overlapping domains \cite{DPINN, cPINN, hpVPINN, XPINN} and adaptive domains \cite{gatedPINN, APINN}. The common denominator across the selection of these methods is that the most challenging part in the domain decomposition for PINN appears to be the computation of the loss function, which determines how well the solution fits at the interface between the sub-domains. 

Another common challenge with domain decomposition approaches is their ability to handle uncertainty in time. For example, approaches like conservative PINN (cPINN) \cite{cPINN} or hp-VPINN \cite{hpVPINN} show good performance in the spatial decomposition of certain PDEs; however, they restrict the flexibility of solving generalized problems and pose additional constraints on problems placed within the time domain\cite{klawonn, cPINN}. On the other hand, other approaches such as XPINN \cite{XPINN} provide a fair generalized framework for solving various types of problems, because they are not bound by conservation laws, which allows for the domain decomposition over time \cite{klawonn, XPINN}. However, the solution of XPINN can be local and local models often experience overfitting as they are trained with only a portion of the available data, adding more uncertainty than vanilla PINN \cite{klawonn, xpinn2}.

Combining Bayesian inference with domain decomposition would enable UQ in cases where subdomains experience different levels of noise. Therefore, we propose \$PINN, a framework that combines the advantages of BPINNs to handle the uncertainties and the advantages of cPINN to work with multi-scale problems. We test our method over different criteria to illustrate the performance and convergence for various problems: different types of PDEs; variable levels of added noise in the data; variable levels of noise in each domain; different domain sizes; forward and inverse models; 1D and 2D multi-scale problems.
\section{Methodology}
This section provides the \$PINN formulation. Initially we introduces the original models that serve as a basis for \$PINN, and later provide the mathematical formulation used in the implementation.
First, we describe the generic composition of PINN, followed by description of the contributions from BPINN and cPINN. The resulting \$PINN architecture is composed of the key building blocks that define BPINN and multi-scale PINN models. 


\subsection{PINNs}
\begin{figure}[h]
    \centering
    \includegraphics[width=0.7\linewidth]{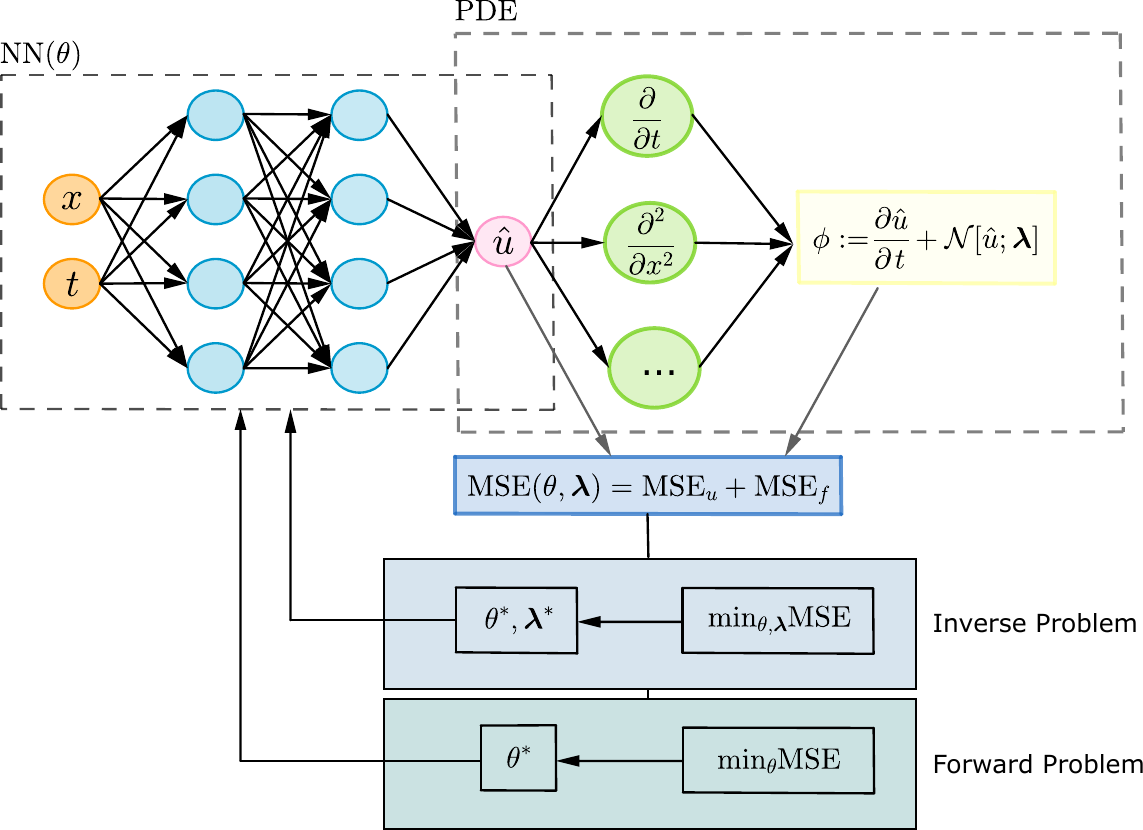}
    \caption{Structure of Vanilla PINN for forward and inverse problems.}
    \label{fig:pinn_forward_inverse}
\end{figure}

PINN is a Scientific Machine Learning (SciML) method for solving differential equations by encoding the physical representation into the neural network. The schematic representation of the general form of PINN is shown in Figure \ref{fig:pinn_forward_inverse}.
We consider a general form of a PDE:
\begin{equation}
    \frac{\partial u}{\partial t} + \mathcal{N}[u;\boldsymbol{\lambda}] = 0, \quad x\in\Omega, \quad t\in[0,n],
    \label{eq:PDE}
\end{equation}
where $\mathcal{N}$ is the differential operator, $u(x,t)$ is the latent hidden solution, $\boldsymbol{\lambda}$ is the vector of the PDE's parameters, $\Omega$ is the space domain, which is a subset of $\mathbb{R}^d$.
For the forward problem, the parameters $\boldsymbol{\lambda}$ are known, so the goal is to find the solution $u(x,t)$ of the PDE. For the inverse problem, the vector of parameters $\boldsymbol{\lambda}$ is inferred from the data. We define the residual as the left-hand side of \eqref{eq:PDE} as follows:
\begin{equation}
    \phi(x,t) =  \frac{\partial u}{\partial t} + \mathcal{N}[u;\boldsymbol{\lambda}].
    \label{eq:general_pde}
\end{equation}

The model is trained by minimizing the loss function defined as the mean-squared error MSE:
\begin{equation}
    \text{MSE} = \text{MSE}_{u} + \text{MSE}_{\phi},
    \label{eq:loss}
\end{equation}
where
\begin{align}
    & \text{MSE}_u = \frac{1}{{N}_u}\sum_{i=1}^{{N}_u}{|\hat{u}(x_u^i, t_u^i)-u^i|^2}, \quad \label{eq:MSE_uf1} \\
    \quad & \text{MSE}_{\phi} = \frac{1}{{N}_f}\sum_{i=1}^{{N}_f}{|\phi(x_{\phi}^i, t_{\phi}^i)|^2},
    \label{eq:MSE_uf}
\end{align}
where $\{x_u^{i},t_u^{i},u^{i}\}_{i=1}^{{N}_u}$ is either the training data given by the initial and boundary conditions in the forward problem formulation or the solution data points across the domain used for training in the inverse problem; $\{x_{\phi}^{i},t_{\phi}^{i}\}_{i=1}^{{N}_{\phi}}$ are the collocation points for $\phi(x,t)$. 


\subsection{BPINN}
\begin{figure}[h]
    \centering    \includegraphics[width=0.7\linewidth]{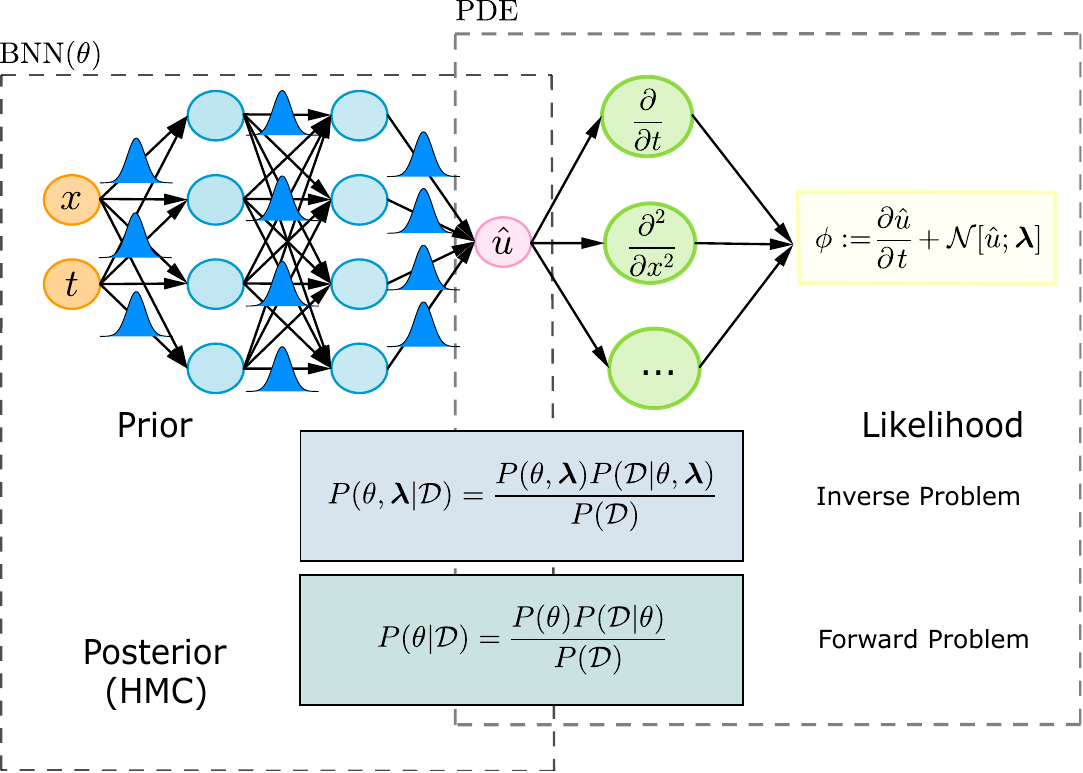}
    \caption{Structure of BPINN for forward and inverse problems.}
    \label{fig:bpinn_forward_inverse}
\end{figure}

BPINNs incorporate Bayesian inference to approximate problems with noisy or unavailable data.
Figure \ref{fig:bpinn_forward_inverse} shows the BPINNs' architecture for the forward and inverse problem. In this case, the first part consists of a BNN approximating the solution $u$. The second part is the physics-informed part, where the residual is defined by \eqref{eq:general_pde}.

We define the dataset $\mathcal{D}$ to be the scattered noisy data of the solution $u$, the residual $\phi$ and the boundary and initial conditions $u_b$ as follows:
\begin{equation}
    \mathcal{D} = \mathcal{D}_u \cup \mathcal{D}_{\phi} \cup \mathcal{D}_b,
\end{equation}
where $\mathcal{D}_u\!=\!\{(x_u, t_u, \overline{u}^{(i)})\}_{i=1}^{N_u}$, $\mathcal{D}_{\phi}\!=\!\{(x_{\phi}, t_{phi}, \overline{\phi}^{(i)})\}_{i=1}^{N_{\phi}}$, $\mathcal{D}_b\!=\!\{(x_b, t_b, \overline{u}_b^{(i)})\}_{i=1}^{N_b}$. 
The data is assumed to be independently Gaussian distributed:
\begin{align}
    & \overline{u}^{(i)} = u(x_u^{(i)}, t_u^{(i)}) + \epsilon_u^{(i)}, \quad i=1,2,\dots,N_u, \\
    & \overline{\phi}^{(i)} = \phi(x_{\phi}^{(i)}, t_{\phi}^{(i)}) + \epsilon_{\phi}^{(i)}, \quad i=1,2,\dots,N_{\phi}, \\
    & \overline{u}_b^{(i)} = u_b(x_b^{(i)}, t_b^{(i)}) + \epsilon_b^{(i)}, \quad i=1,2,\dots,N_b, 
\end{align}
where $\epsilon_u^{(i)}$, $\epsilon_{\phi}^{(i)}$, and $\epsilon_b^{(i)}$ are independent Gaussian noises with zero mean. The corresponding standard deviations $\sigma_u^{(i)}$, $\sigma_{\phi}^{(i)}$, and $\sigma_b^{(i)}$ are known. 
First, we consider the forward problem formulation. The Bayesian framework is represented considering a surrogate model $\hat{u}$. The likelihood is defined as:
\begin{equation}
    P(\mathcal{D}\! \mid \! \boldsymbol{\theta}) = P(\mathcal{D}_u\! \mid \! \boldsymbol{\theta})P(\mathcal{D}_{\phi}\! \mid \! \boldsymbol{\theta})P(\mathcal{D}_{b}\! \mid \! \boldsymbol{\theta}),
    \label{eq:bpinn_all}
\end{equation}
where
\begin{align}
    & P (\mathcal{D}_{u} \! \mid \! \boldsymbol{\theta}) = \!\prod_{i=1}^{N_{u}} \!  \frac{1}{\sqrt{2 \pi \sigma_u^{(i)^2}}} \exp \! \! \left[\! -\frac{\left({\hat{u}}(x_u^{(i)}, t_u^{(i)} ; \boldsymbol{\theta})-\overline{u}^{(i)}\right)^2}{2 \sigma_u^{(i)^2}}\right]\!\!, \label{eq:bpinn1}\\ 
     & P (\mathcal{D}_{\phi} \! \mid \! \boldsymbol{\theta}) = \!\prod_{i=1}^{N_{\phi}} \!  \frac{1}{\sqrt{2 \pi \sigma_{\phi}^{(i)^2}}} \exp \! \! \left[\! -\frac{\left({\phi}(x_{\phi}^{(i)}, t_{\phi}^{(i)} ; \boldsymbol{\theta})-\overline{\phi}^{(i)}\right)^2}{2 \sigma_{\phi}^{(i)^2}}\right]\!\!, \label{eq:bpinn2}
     \end{align}
     \begin{align}
     & P (\mathcal{D}_{b} \! \mid \! \boldsymbol{\theta}) = \!\prod_{i=1}^{N_{b}} \!  \frac{1}{\sqrt{2 \pi \sigma_{b}^{(i)^2}}} \exp \! \! \left[\! -\frac{\left({\hat{u}_b}(x_{b}^{(i)}, t_{b}^{(i)} ; \boldsymbol{\theta})-\overline{u}_b^{(i)}\right)^2}{2 \sigma_b^{(i)^2}}\right]\!\!. \label{eq:bpinn3}
\end{align}
The posterior is retrieved using Bayes' theorem:
\begin{equation}
    P(\boldsymbol{\theta}\! \mid \! \mathcal{D}) = \frac{ P(\mathcal{D}\! \mid \! \boldsymbol{\theta})P(\boldsymbol{\theta})}{P(\mathcal{D})} \simeq P(\mathcal{D}\! \mid \! \boldsymbol{\theta})P(\boldsymbol{\theta}).
\end{equation}
For the inverse problem, the formulation is very similar to the above. However, we additionally assign a prior distribution to the unknown parameters $\boldsymbol{\lambda}$. The joint posterior for [$\boldsymbol{\theta}$,$\boldsymbol{\lambda}$] is as follows:
\begin{equation}
    P(\boldsymbol{\theta}\!, \!\boldsymbol{\lambda}\!\! \mid \! \mathcal{D}) \!=\! \frac{ P(\mathcal{D}\!\! \mid \! \boldsymbol{\theta}\!,\!\boldsymbol{\lambda})P(\boldsymbol{\theta}\!,\!\boldsymbol{\lambda})}{P(\mathcal{D})} \!\simeq \!P(\mathcal{D}\!\! \mid \! \boldsymbol{\theta}\!,\!\boldsymbol{\lambda})P(\boldsymbol{\theta}\!,\!\boldsymbol{\lambda}) \!= \!P(\mathcal{D}\!\! \mid \! \boldsymbol{\theta}\!,\!\boldsymbol{\lambda})P(\boldsymbol{\theta})P(\boldsymbol{\lambda}).
\end{equation}



\subsection{cPINN}
\begin{figure}[h]
    \centering
    \includegraphics[width=1.0\linewidth]{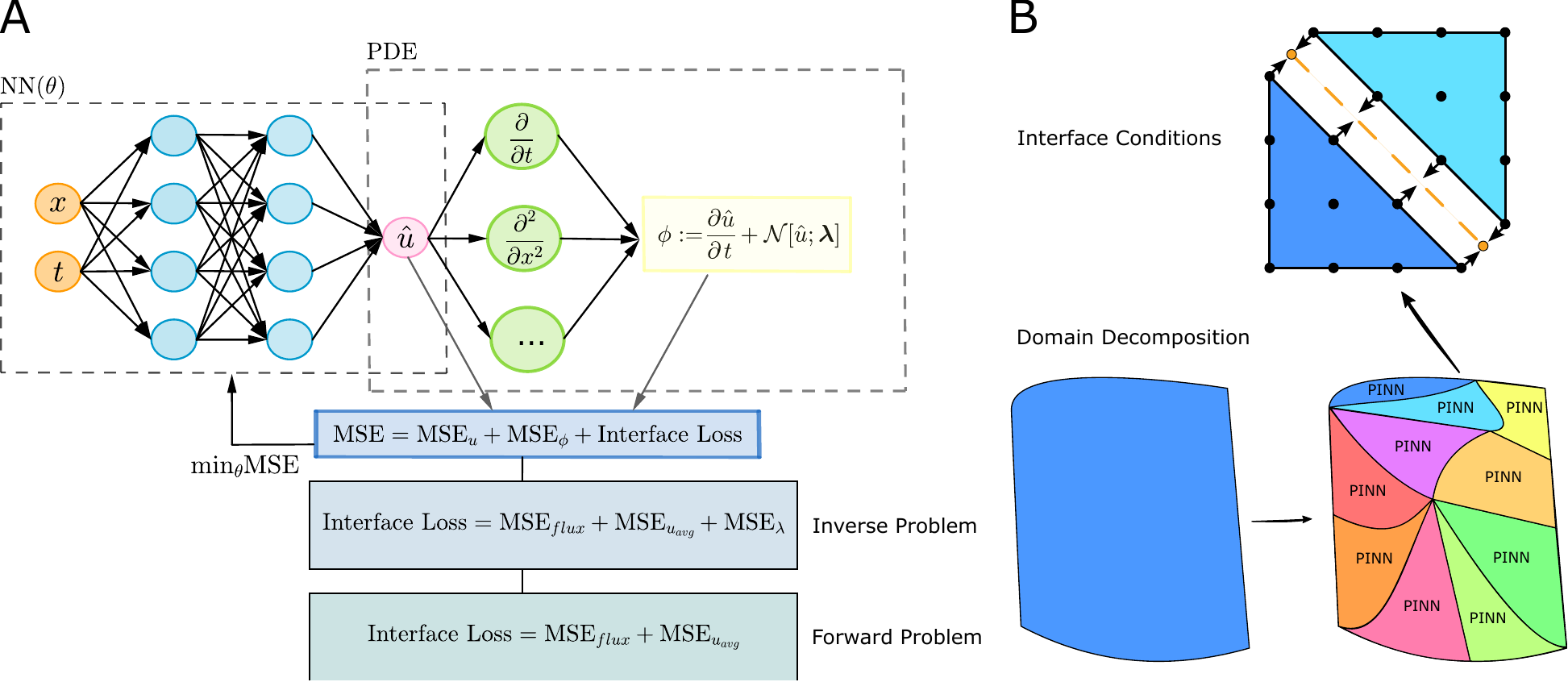}
    \caption{Structure of cPINN for forward and inverse problems.}
    \label{fig:cpinn_forward_inverse}
\end{figure}

cPINN is a domain decomposition method that trains networks to obtain solutions for each subdomain, considering the interface conditions between the subdomains. This is done by adding an extra interface term to the loss function, which would compute and minimize the error at each interface. 
Figure \ref{fig:cpinn_forward_inverse}, on the left (A), shows the schematic representation of cPINN for the forward and inverse problems, and, on the right (B), for domain decomposition and interface conditions. Domain decomposition ensures that the flux at the interface is imposed in the normal direction.  
Given the residual as defined in \eqref{eq:general_pde}, the cPINN loss function for the forward problem can be defined as:

\begin{equation}
    \text{MSE} = \text{MSE}_{u} + \text{MSE}_{\phi} + \text{Interface Loss},
    \label{eq:losscPINN}
\end{equation}
where 
\begin{align}
& \text{Interface Loss} = \text{MSE}_{u_{avg}} + \text{MSE}_{flux}, \\
    & \text{MSE}_{u_{avg}} = \frac{1}{N_{I_q}}\sum_{i=1}^{{N}_{I_q}}|\hat{u_{q}}(x_{I_q}^i,t_{I_q}^i)-\{ \{ u(x_{I_q}^i,t_{I_q}^i)\}\}|^2, \\
    & \text{MSE}_{flux} = \frac{1}{N_{I_q}}\sum_{i=1}^{{N}_{I_q}}{|f_q(u(x_{I_q}^i,t_{I_q}^i))\cdot \textbf{n} - f_{q^+}(u(x_{I_q}^{i},t_{I_q}^{i}))\cdot \textbf{n})|^2},
\end{align}
where $f_p\cdot \textbf{n}$ and $f_{q^+}\cdot \textbf{n}$ are the interface fluxes normal to the common interface of two subdomains $q$ and $q^+$, respectively. Indeed, the superscript $+$ represents the bordering subdomain. $N_{I_q}$ is the number of points on the common interface in the $q$-th subdomain. The subscript $I_q$ relates to the interface between subdomains $q$ and $q^+$, so $x_{I_q}$ and $t_{I_q}$ are coordinates of common interface points. $\{\{u\}\}$ is the average value of $u$ at the common interface and is given by:
\begin{equation}
    \{\{u\}\} = u_{avg} := \frac{u_q + u_{q^+}}{2}.
\end{equation}

For the inverse problem, the formulation is similar; however, in the interface loss, we have an additional loss term for the unknown parameters $\boldsymbol{\lambda}$ that guarantees continuity condition on the interface, expressed as:

\begin{equation}
    \text{MSE}_{\boldsymbol{\lambda}} = \frac{1}{N_{I_q}}\sum_{i=1}^{{N}_{I_q}}|\boldsymbol{\lambda}_q(x_{I_q}^i,t_{I_q}^i)-\boldsymbol{\lambda}_{q^{+}}(x_{I_q}^i,t_{I_q}^i)|^2.
\end{equation}



\subsection{\$PINN}

\begin{figure}[ht]
    \centering
    \includegraphics[width=0.8\linewidth]{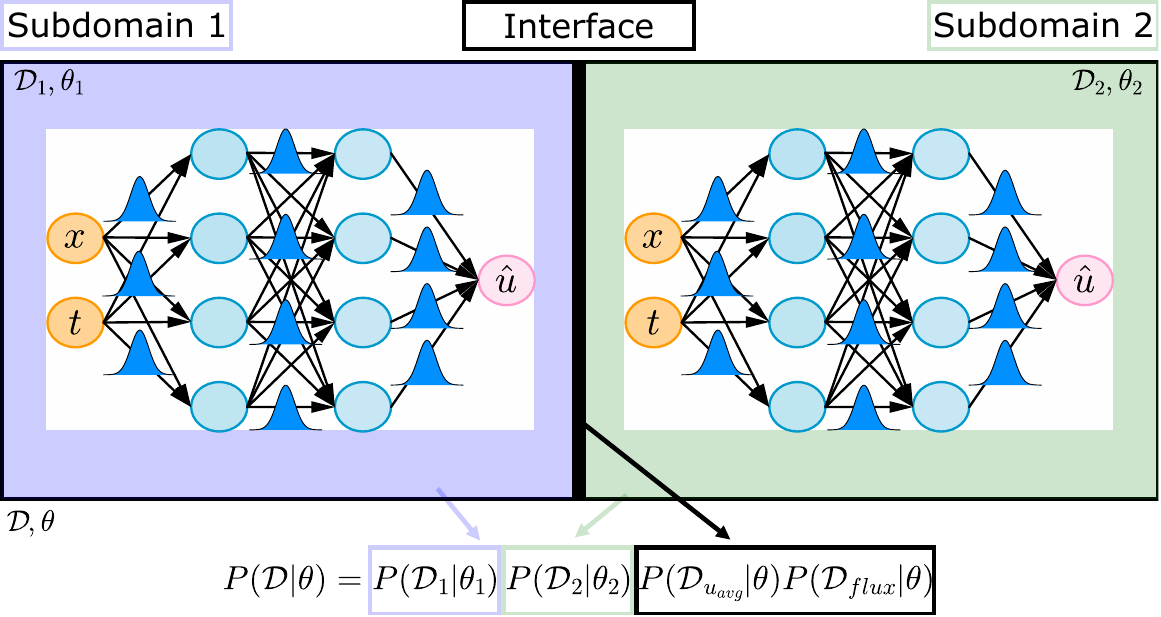}
    \caption{Structure of \$PINN.}
    \label{fig:dollarpinn}
\end{figure}

The proposed \$PINN architecture is presented in Figure \ref{fig:dollarpinn}, where a domain split into two subdomains with the shared interface is shown. 

The mathematical formulation of the likelihoods is the following:
\begin{equation}
    P(\mathcal{D}\! \mid \! \boldsymbol{\theta}) \!=\! P(\mathcal{D}_u\!\! \mid \!\! \boldsymbol{\theta})P(\mathcal{D}_{\phi}\!\! \mid \!\! \boldsymbol{\theta})P(\mathcal{D}_{IC}\!\! \mid \!\! \boldsymbol{\theta})P(\mathcal{D}_{BC}\!\! \mid \!\! \boldsymbol{\theta})P(\mathcal{D}_{u_{avg}}\!\! \mid \!\! \boldsymbol{\theta})P(\mathcal{D}_{flux}\!\! \mid \!\! \boldsymbol{\theta}),
    \label{eq:spinn_all}
\end{equation}
where
\begin{align}
    & P (\mathcal{D}_{u} \! \mid \! \boldsymbol{\theta}) = \!\prod_{i=1}^{N_{u}} \!  \frac{1}{\sqrt{2 \pi \sigma_u^{(i)^2}}} \exp \! \! \left[\! -\frac{\left({\hat{u}}(x_u^{(i)}, t_u^{(i)} ; \boldsymbol{\theta})-u^{(i)}\right)^2}{2 \sigma_u^{(i)^2}}\right]\!\!, \label{eq:spinn1}\\ 
     & P (\mathcal{D}_{\phi} \! \mid \! \boldsymbol{\theta}) = \!\prod_{i=1}^{N_{\phi}} \!  \frac{1}{\sqrt{2 \pi \sigma_{\phi}^{(i)^2}}} \exp \! \! \left[\! -\frac{\left({\phi}(x_{\phi}^{(i)}, t_{\phi}^{(i)} ; \boldsymbol{\theta})-\phi^{(i)}\right)^2}{2 \sigma_{\phi}^{(i)^2}}\right]\!\!, \label{eq:spinn2}\\
     & P (\mathcal{D}_{IC} \!\! \mid \!\! \boldsymbol{\theta}) = \!\prod_{i=1}^{N_{IC}} \!  \frac{1}{\sqrt{2 \pi \sigma_{IC}^{(i)^2}}} \exp \! \! \left[\! -\frac{\left({\hat{u}}(x_{IC}^{(i)}, 0 ; \boldsymbol{\theta})-u(x_{IC}^{(i)}, 0 )\right)^2}{2 \sigma_{IC}^{(i)^2}}\right]\!\!, \label{eq:spinn3}\\ 
     & P (\mathcal{D}_{BC} \! \mid \! \boldsymbol{\theta}) = \!\prod_{i=1}^{N_{BC}} \!  \frac{1}{\sqrt{2 \pi \sigma_{BC}^{(i)^2}}} \exp \! \! \left[\! -\frac{\left({\hat{u}}(x_{BC}^{(i)}, t_{BC}^{(i)} ; \boldsymbol{\theta})-u^{(i)}\right)^2}{2 \sigma_{BC}^{(i)^2}}\right]\!\!, \label{eq:spinn4}\\ 
     & P (\mathcal{D}_{u_{avg}} \! \mid \! \boldsymbol{\theta}) = \!\prod_{i=1}^{N_{CDC}} \!  \frac{1}{\sqrt{2 \pi \sigma_{u_{avg}}^{(i)^2}}} \exp \! \! \left[\! -\frac{\left(\! \mid \!{\hat{u}_1}(x_{CDC}^{(i)}, t_{CDC}^{(i)} ; \boldsymbol{\theta})-\mathcal{I}_{u_{avg}}\! \mid \!\right)^2}{2 \sigma_{u_{avg}}^{(i)^2}}\right]\!\!, \label{eq:spinn5} \\ 
     & P (\mathcal{D}_{flux} \!\! \mid \!\! \boldsymbol{\theta}) \!\!= \!\!\! \prod_{i=1}^{N_{CDC}} \!\!\!  \frac{1}{\sqrt{2 \pi \sigma_{flux}^{(i)^2}}} \! \exp \!\!\! \left[\! -\frac{\left(\! \mid \!f_1({\hat{u}_1}(x_{CDC}^{(i)}, t_{CDC}^{(i)} ; \boldsymbol{\theta}))\! \cdot\! \mathbf{n}\!-\! \mathcal{I}_{flux}\! \mid \!\right)^2}{2 \sigma_{flux}^{(i)^2}}\right]\!\!, \label{eq:spinn6}
\end{align}
where $P (\mathcal{D}_{u} \! \mid \! \boldsymbol{\theta})$ is the data likelihood for observed data; $P (\mathcal{D}_{\phi} \! \mid \! \boldsymbol{\theta})$ is the residual likelihood enforcing the constraints imposed by the PDE; $P (\mathcal{D}_{IC} \! \mid \! \boldsymbol{\theta})$ and $P (\mathcal{D}_{BC} \! \mid \! \boldsymbol{\theta})$ are the likelihoods for the initial and boundary data, respectively; $P (\mathcal{D}_{u_{avg}} \! \mid \! \boldsymbol{\theta})$ and $P (\mathcal{D}_{flux} \! \mid \! \boldsymbol{\theta})$ are the interface likelihoods, the key contribution of our model. These additional constraints ensure the continuity of both solution values and flow at the interface. The subscript $CDC$ stands for common domain conditions or collocation points at the interface. 
In \eqref{eq:spinn5} and \eqref{eq:spinn6}, the terms $\mathcal{I}_{u_{avg}}$ and $\mathcal{I}_{flux}$ are defined as:
\begin{align}
    & \mathcal{I}_{u_{avg}}=\frac{\hat{u}_1(x_{CDC}^{(i)}, t_{CDC}^{(i)} ; \boldsymbol{\theta})+\hat{u}_2(x_{CDC}^{(i)}, t_{CDC}^{(i)} ; \boldsymbol{\theta})}{2}, \\
    & \mathcal{I}_{flux}=\frac{f_1(\hat{u}_1(x_{CDC}^{(i)}, t_{CDC}^{(i)} ; \boldsymbol{\theta}))\cdot \boldsymbol{n} +f_2(\hat{u}_2(x_{CDC}^{(i)}, t_{CDC}^{(i)} ; \boldsymbol{\theta}))\cdot \boldsymbol{n}}{2}.
\end{align}

For the inverse problem, we have a similar setup for the likelihoods as in~\eqref{eq:spinn_all}. However, the residual likelihood changes by conditioning the probability distribution on the parameters we want to estimate $\boldsymbol{\lambda}$. In this way, we ensure that the model trains while sampling the unknown parameters from posterior. In particular, if we consider two subdomains, we will split the vector parameters into $\boldsymbol{\lambda}_1$ and $\boldsymbol{\lambda}_2$, one for each subdomain. Therefore, the residual likelihood becomes:
\begin{equation} \label{inverse1}
    P (\mathcal{D}_{\phi} \! \mid \! \boldsymbol{\theta}; \boldsymbol{\lambda}) = P (\mathcal{D}_{\phi_1} \! \mid \! \boldsymbol{\theta}; \boldsymbol{\lambda}_1)P (\mathcal{D}_{\phi_2} \! \mid \! \boldsymbol{\theta}; \boldsymbol{\lambda}_2),
\end{equation}

where
\begin{align}
     & P (\mathcal{D}_{\phi_1} \! \mid \! \boldsymbol{\theta}; \boldsymbol{\lambda}_1) = \!\prod_{i=1}^{N_{\phi}} \!  \frac{1}{\sqrt{2 \pi \sigma_{\phi_1}^{(i)^2}}} \exp \! \! \left[\! -\frac{\left({\phi}(x_{\phi_1}^{(i)}, t_{\phi_1}^{(i)} ; \boldsymbol{\theta}; \boldsymbol{\lambda}_1)-\phi^{(i)}\right)^2}{2 \sigma_{\phi_1}^{(i)^2}}\right]\!\!,
     \label{eq:spinn_inv1}\\
     & P (\mathcal{D}_{\phi_2} \! \mid \! \boldsymbol{\theta}; \boldsymbol{\lambda}_2) = \!\prod_{i=1}^{N_{\phi}} \!  \frac{1}{\sqrt{2 \pi \sigma_{\phi_2}^{(i)^2}}} \exp \! \! \left[\! -\frac{\left({\phi}(x_{\phi_2}^{(i)}, t_{\phi_1}^{(i)} ; \boldsymbol{\theta}; \boldsymbol{\lambda}_2)-\phi^{(i)}\right)^2}{2 \sigma_{\phi_2}^{(i)^2}}\right]\!\!. \label{eq:spinn_inv2}
\end{align}

The subscripts $\phi_1$ and $\phi_2$ stand for the residual $\phi$ evaluated at the subdomains 1 and 2, respectively.  
As this setup could create possible discrepancies between the predictions of the unknown parameters in the two domains, we introduce two types of constraints: a soft and a hard constraint. The soft condition implies that both $\boldsymbol{\lambda}_1$ and $\boldsymbol{\lambda}_2$ tend to be equal and is expressed as follows:
\begin{equation}
    P (\mathcal{D}_{\lambda} \! \mid \! \boldsymbol{\theta}) = \!\prod_{i=1}^{N_{CDC}} \!  \frac{1}{\sqrt{2 \pi \sigma_{\lambda}^{(i)^2}}} \exp \! \! \left[\! -\frac{\left(\boldsymbol{\lambda}_1 - \frac{\boldsymbol{\lambda}_1+\boldsymbol{\lambda}_2}{2}\right)^2}{2 \sigma_{\lambda}^{(i)^2}}\right]\!\! .
    \label{eq:soft_cond}
\end{equation}

However, in this way the parameters  $\boldsymbol{\lambda}_1$ and $\boldsymbol{\lambda}_2$ are still estimated using the two different residual likelihoods, and therefore, they differ in the two domains. 
For this reason, we also test the model using a hard constraint. Instead of having one vector parameter for each domain, we define only one $\boldsymbol{\lambda}$ for both subdomains, which will be learned using the two BPINNs defined in each subdomain. Therefore, hard condition implies: 
\begin{equation}
    P (\mathcal{D}_{\phi} \! \mid \! \boldsymbol{\theta}; \boldsymbol{\lambda}) = P (\mathcal{D}_{\phi_1} \! \mid \! \boldsymbol{\theta}; \boldsymbol{\lambda})P (\mathcal{D}_{\phi_2} \! \mid \! \boldsymbol{\theta}; \boldsymbol{\lambda}),
    \label{eq:hard_cond}
\end{equation}
where
\begin{align} 
     & P (\mathcal{D}_{\phi_1} \! \mid \! \boldsymbol{\theta}; \boldsymbol{\lambda}) = \!\prod_{i=1}^{N_{\phi}} \!  \frac{1}{\sqrt{2 \pi \sigma_{\phi_1}^{(i)^2}}} \exp \! \! \left[\! -\frac{\left({\phi}(x_{\phi_1}^{(i)}, t_{\phi_1}^{(i)} ; \boldsymbol{\theta}; \boldsymbol{\lambda})-\phi^{(i)}\right)^2}{2 \sigma_{\phi_1}^{(i)^2}}\right],\!\! \label{eq:spinn_inv1_hc}\\
     & P (\mathcal{D}_{\phi_2} \! \mid \! \boldsymbol{\theta}; \boldsymbol{\lambda}) = \!\prod_{i=1}^{N_{\phi}} \!  \frac{1}{\sqrt{2 \pi \sigma_{\phi_2}^{(i)^2}}} \exp \! \! \left[\! -\frac{\left({\phi}(x_{\phi_2}^{(i)}, t_{\phi_1}^{(i)} ; \boldsymbol{\theta}; \boldsymbol{\lambda})-\phi^{(i)}\right)^2}{2 \sigma_{\phi_2}^{(i)^2}}\right].\!\! \label{eq:spinn_inv2_hc}
\end{align}

\section{Model Implementation}
In order to illustrate the performance of \$PINN we run a series of computational experiments using 4 different PDEs: Burgers' equation, Fisher-KPP equation, Fokker-Planck equation and Allen-Cahn equation. Additionally, we summarize the performed computational experiments and outline the implementation strategy for both forwards and inverse problems with different levels of added noise and variable domain sizes.  

\subsection{Analyzed PDEs}
\label{subsec:analyzed_pdes}
This section outlines the PDEs used in testing the proposed method. The choice of the equations is motivated by their complexity and type since PINN often shows different performance and convergence rates depending on the type of the PDE. 

\paragraph{\textbf{Burgers' equation}}

The general form of viscous Burgers' equation in 1D spatio-temporal domain is given by:
\begin{equation}
    \frac{\partial u}{\partial t} + u\frac{\partial u}{\partial x}= \nu \frac{\partial^2 u}{\partial x^2},
    \label{Burg1}
\end{equation}
where $\nu$ is the diffusion coefficient, $\nu=0.01/\pi$. We consider the spatio-temporal domain $(x,t)\in [-1,1]\times[0,1]$ with boundary and initial conditions specified as:
\begin{align}
    & u(t, -1) = u(t, 1) = 0, \label{Burg3} \\
    & u(0, x) = -\sin(\pi x). \label{Burg4}    
\end{align}


\paragraph{\textbf{Fisher-KPP equation}} Fisher-KPP is a diffusion-reaction equation often used to model wave propagation or population growth. The general form of Fisher-KPP PDE is expressed in:
\begin{equation}
    \frac{\partial u}{\partial t} - D\frac{\partial^2 u}{\partial x^2} = ru (1-u),
    \label{FKPP}
\end{equation}
where $D$ is the diffusion constant, $D=0.1$; $r$ is the growth rate, $r=2$; and $f(u,x,t)=ru(1-u)$ represents an inhomogeneous term. We consider the spatio-temporal domain $(x,t)\in [-1,1]\times[0,1]$ with boundary and initial conditions specified in \eqref{FKPP3}-\eqref{FKPP4}.

\begin{align}
    & u(t, -1) = u(t, 1) = 0, \label{FKPP3}\\
    & u(0, x) = e^{-x^2}. 
    \label{FKPP4}
\end{align}
\paragraph{\textbf{Fokker-Planck equation}}
The Fokker-Planck equation is commonly used to describe the time evolution of the particle velocity caused by drag and random forces as a probability density function. The general form of the Fokker-Planck equation is:
\begin{equation}
    \frac{\partial u}{\partial t} - D \frac{\partial^2 u}{\partial x^2} = 0,
    \label{FP1}
\end{equation}
where $D$ is the diffusion constant, $D=0.1$. We consider the spatio-temporal domain $(x,t)\in [-1,1]\times[0,1]$ with boundary and initial conditions specified in \eqref{FP3}-\eqref{FP4}.
\begin{align}
    & u(t, -1) = u(t, 1) = 0, \label{FP3} \\
    & u(0, x) = \frac{1}{\sqrt{2\pi\sigma^2}} e^{-0.5x^2\sigma^2},
    \label{FP4}
\end{align}
where $D$ is a diffusion constant, $D=0.1$; $\sigma$ is the noise ratio $\sigma=0.2$.

\paragraph{\textbf{Allen-Cahn equation}}
The Allen-Cahn equation is commonly used for modeling multiphase flows. This study considers a diffusion reaction in porous media, with a general form described in \eqref{AC}. 
\begin{equation}
    \frac{\partial u}{\partial t} - D\frac{\partial^2 u}{\partial x^2} + \lambda(x) u^3 - f(x) = 0,
    \label{AC}
\end{equation}
where $D$ is the diffusion constant, $D=0.01$, ; $\lambda(x)$ represents space dependent mobility or reaction rate; $f(x)$ is the source term; $t$ and $x$ are the time and space coordinates; $u(t,x)$ is the concentration. We consider the spatio-temporal domain $(x,t)\in [-1,1]\times[0,1]$, $l=0.4$ with boundary and initial conditions specified in \eqref{AC2}-\eqref{AC5}.
\begin{align}
    & u(t, -1) = u(t, 1) = 0.5, \label{AC2} \\
    & u(0, x) = 0.5 \cos^2(\pi x), \label{AC3} \\
    & \lambda(x) = 0.2 + e^x\cos^2(2x),
    \label{AC4}\\
    &f(x) = e^{-\frac{(x-0.25)^2}{2l^2}}\sin^2(3x).
    \label{AC5}
\end{align}


\subsection{Computational experiments}
\label{sec:computational_experiments}
\begin{figure}[ht]
     \centering
     \begin{subfigure}[b]{0.49\textwidth}
         \centering
         \includegraphics[width=\textwidth]{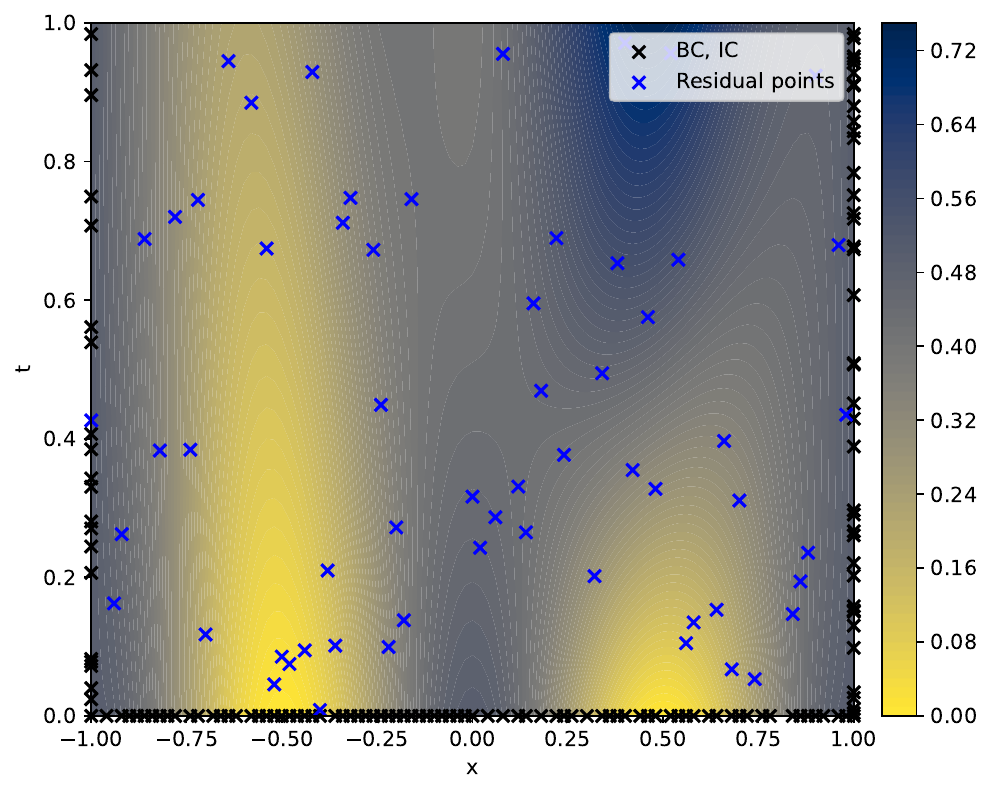}
         \caption{BI}
         \label{fig:train_bpinn_bi}
     \end{subfigure}
     \begin{subfigure}[b]{0.49\textwidth}
         \centering
         \includegraphics[width=\textwidth]{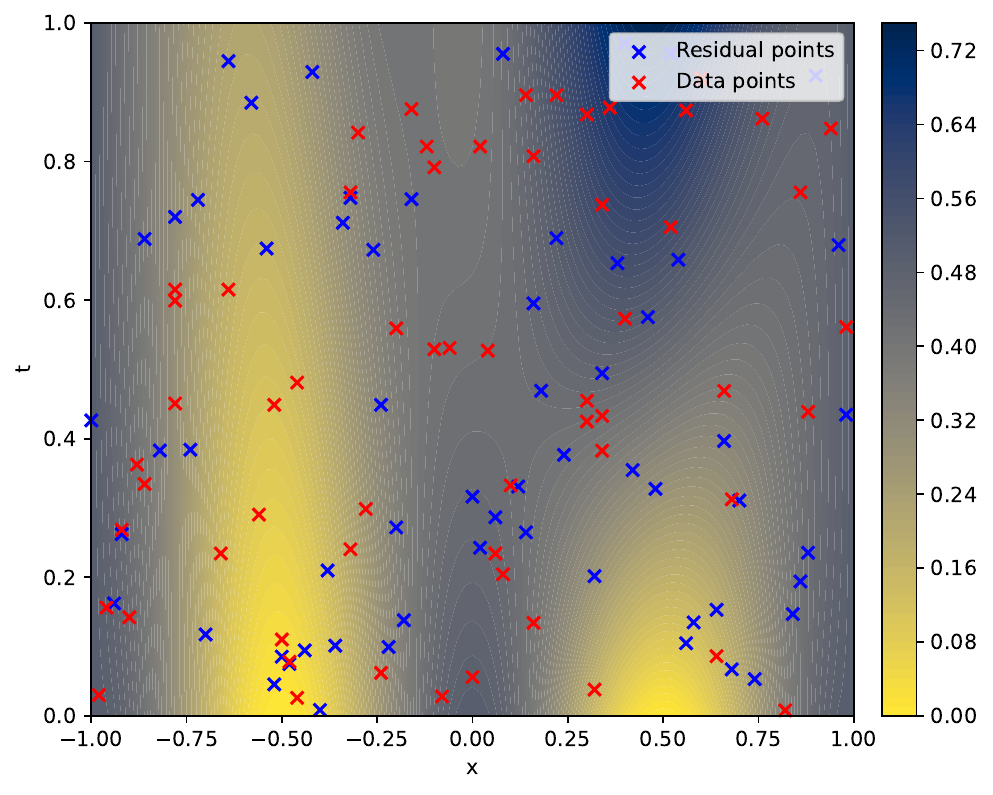}
         \caption{RD}
         \label{fig:train_bpinn_rd}
     \end{subfigure}
     \caption{Sampling of training data points for BPINN for BI and RD.}
        \label{fig:train_bpinn_all}
\end{figure}

We have tested two different training strategies that mainly differ in the input data used to inform PINN during training. Figure \ref{fig:train_bpinn_all} shows an example of both approaches for BPINN. In the first scenario, Figure \ref{fig:train_bpinn_bi}, for training we use a combination of initial and boundary conditions, as well as residual or collocation points. We will refer to this scenario as BI. The second approach shown in Figure \ref{fig:train_bpinn_rd} abandons initial and boundary conditions in favor of randomly sampled data points throughout the domain and trains the model using the provided data and sampled collocation points and is defined as RD.

\begin{figure}[H]
     \centering
     \begin{subfigure}[b]{0.49\textwidth}
         \centering
         \includegraphics[width=\textwidth]{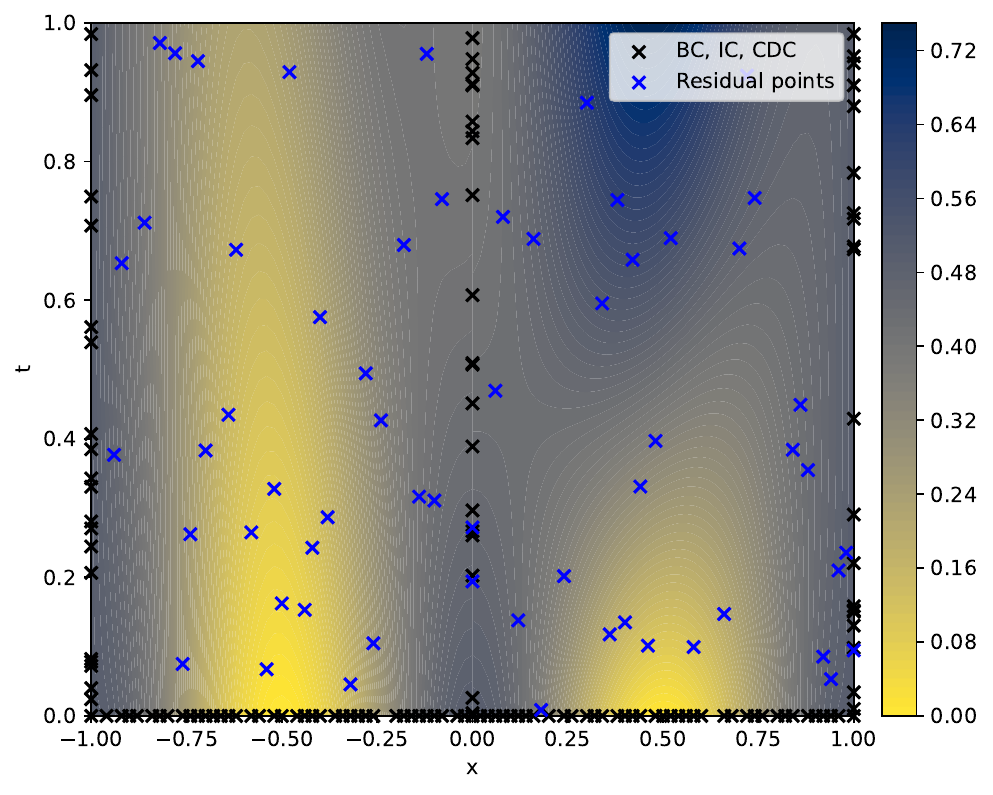}
         \caption{BI}
         \label{fig:train_dollpinn_bi}
     \end{subfigure}
     \begin{subfigure}[b]{0.49\textwidth}
         \centering
         \includegraphics[width=\textwidth]{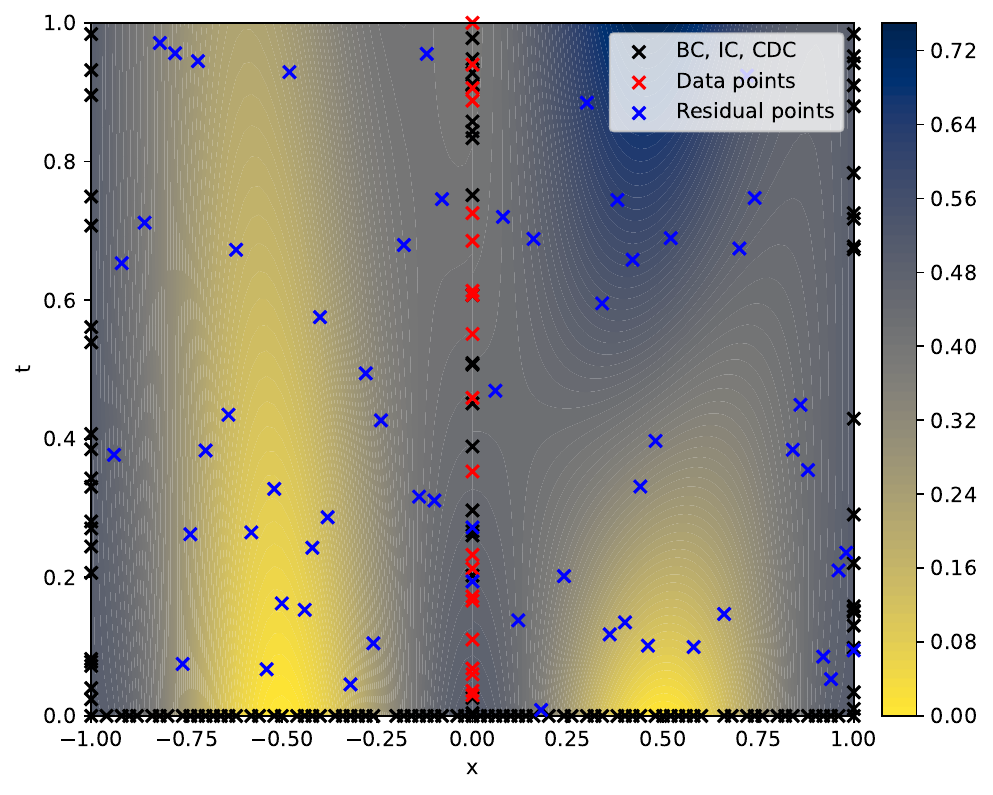}
         \caption{BIC}
         \label{fig:train_dollpinn_bic}
     \end{subfigure}
     \begin{subfigure}[b]{0.49\textwidth}
         \centering
         \includegraphics[width=\textwidth]{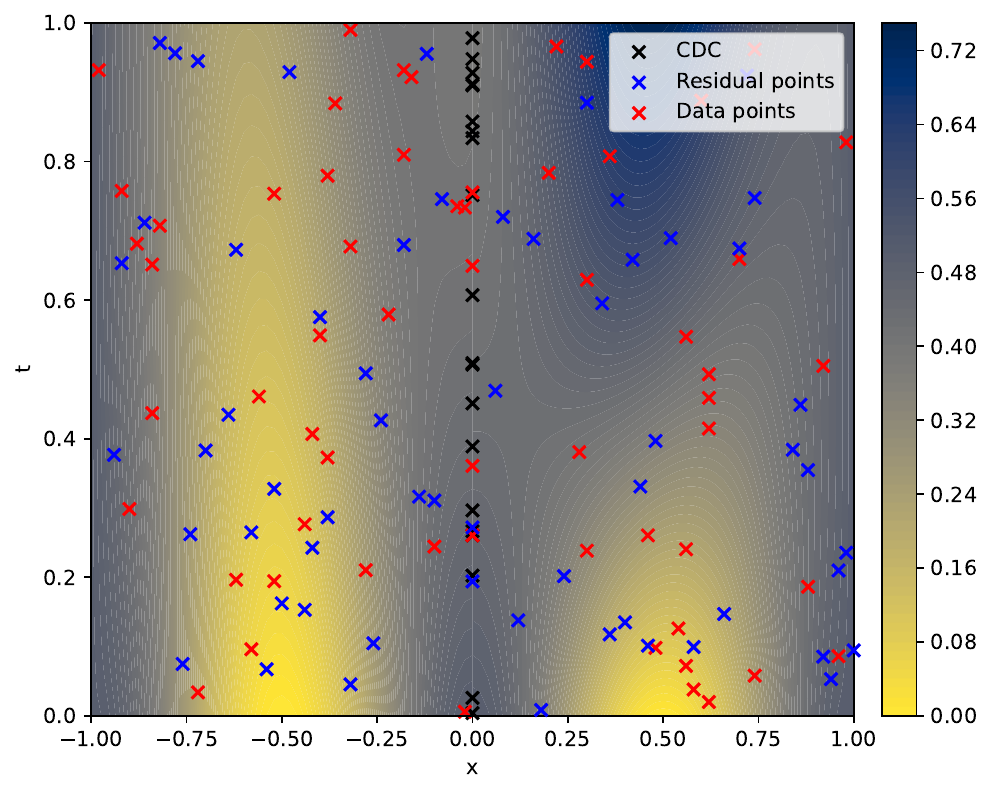}
         \caption{RD}
         \label{fig:train_dollpinn_rd}
     \end{subfigure}
     \caption{Sampling of training data points for \$PINN for BI, BIC and RD.}
        \label{fig:train_dollpinn_all}
\end{figure}

For the \$PINN implementation, an additional set of points along the domain interface is introduced, as shown in Figure \ref{fig:train_dollpinn_bi}, which helps to evaluate and enforce continuity through the likelihood term. 
To identify the main source of uncertainty at the interface, a sub-case is considered in which data points are also provided at the interface, see Figure \ref{fig:train_dollpinn_bic}. In this way, if the uncertainty decreases for BIC sub-case compared to the BI case, it can be inferred that the primary source of uncertainty stems from the lack of data rather than the domain decomposition.
Similarly to the RD problem for BPINN, Figure \ref{fig:train_dollpinn_rd} illustrates that the residual points and the interface conditions are used along with the randomly sampled data points across the domain. 

The following is the summary of three training scenarios shown in Figures \ref{fig:train_bpinn_all} and \ref{fig:train_dollpinn_all}, the results of this study include a comparison between BPINN and \$PINN scenarios for BI and RD training, with an additional case of BIC obtained for \$PINN:
\begin{itemize}
    \item BI - We provide information of the PDE with noisy data of $f(x)$, continuity of the concentration $u_1-(u_1+u_2)/2$ and the flux $\frac{\partial u_1}{\partial x}-(\frac{\partial u_1}{\partial x}+\frac{\partial u_2}{\partial x})/2$ with noise. We provide information about the initial and boundary conditions with noise. 
\item BIC - We provide information of the PDE with noisy data of $f(x)$, continuity of the concentration $u_1-(u_1+u_2)/2$ and the flux $\frac{\partial u_1}{\partial x}-(\frac{\partial u_1}{\partial x}+\frac{\partial u_2}{\partial x})/2$ with noise. We provide information about the initial and boundary conditions with noise. Additionally, in that case, we provide some random values of the data of the co-domain.
\item RD - We provide information of the PDE with noisy data of $f(x)$, continuity of the concentration $u_1-(u_1+u_2)/2$ and the flux $\frac{\partial u_1}{\partial x}-(\frac{\partial u_1}{\partial x}+\frac{\partial u_2}{\partial x})/2$ with noise. Additionally, in that case, we provide some random data values.
\end{itemize}

To ensure robustness and replicability of the \$PINN we also perform test with variable levels of noise and domain size for different PDEs. The performed computational experiments 
contain both forward and inverse problem implementations as well as 1D and 2D case for selected PDEs. Table \ref{tab1} summarizes all tests performed on different PDEs for \$PINN. The summary of the notations for different computation experiments is listed below: 

\begin{itemize}

\item DN - Variable levels of added noise. Gaussian noise is added to the input data, allowing us to test the robustness of \$PINN to higher levels of uncertainty in the input data; the noise level range used is 0-15\% with a 5\% step increase.
\item DS - Different domain sizes have been used to test the model performance for domains of uneven size and see how robust it is to changing interface conditions. 
\item DNS - Combination test of variable noise levels and variable domain sizes. 
\item FP \& IP - Forward problem (FP) and Inverse problem (IP).
\end{itemize}


\begin{table}[ht]
\caption{Tests performed on different PDEs using BPINN with domain decomposition}
\resizebox{\textwidth}{!}{
\begin{tabular}{|lll|llllllllllllllll|}
\hline
\multicolumn{3}{|l|}{\multirow{3}{*}{Case studies}} & \multicolumn{4}{l|}{Fokker-Planck} & \multicolumn{4}{l|}{Fisher-KPP} & \multicolumn{4}{l|}{Burgers'} & \multicolumn{4}{l|}{Allen-Cahn} \\ \cline{4-19} 
\multicolumn{3}{|l|}{} & \multicolumn{16}{c|}{Noise level, \%} \\ \cline{4-19} 
\multicolumn{3}{|l|}{} & \multicolumn{1}{l|}{0} & \multicolumn{1}{l|}{5} & \multicolumn{1}{l|}{10} & \multicolumn{1}{l|}{15} & \multicolumn{1}{l|}{0} & \multicolumn{1}{l|}{5} & \multicolumn{1}{l|}{10} & \multicolumn{1}{l|}{15} & \multicolumn{1}{l|}{0} & \multicolumn{1}{l|}{5} & \multicolumn{1}{l|}{10} & \multicolumn{1}{l|}{15} & \multicolumn{1}{l|}{0} & \multicolumn{1}{l|}{5} & \multicolumn{1}{l|}{10} & 15 \\ \hline
\multicolumn{1}{|l|}{\multirow{12}{*}{FP 1D}} & \multicolumn{1}{l|}{\multirow{3}{*}{}} & BI & \multicolumn{1}{l|}{x} & \multicolumn{1}{l|}{x} & \multicolumn{1}{l|}{x} & \multicolumn{1}{l|}{x} & \multicolumn{1}{l|}{x} & \multicolumn{1}{l|}{x} & \multicolumn{1}{l|}{x} & \multicolumn{1}{l|}{x} & \multicolumn{1}{l|}{x} & \multicolumn{1}{l|}{x} & \multicolumn{1}{l|}{x} & \multicolumn{1}{l|}{x} & \multicolumn{1}{l|}{x} & \multicolumn{1}{l|}{x} & \multicolumn{1}{l|}{x} & x \\ \cline{3-19} 
\multicolumn{1}{|l|}{} & \multicolumn{1}{l|}{} & BIC & \multicolumn{1}{l|}{x} & \multicolumn{1}{l|}{x} & \multicolumn{1}{l|}{x} & \multicolumn{1}{l|}{x} & \multicolumn{1}{l|}{x} & \multicolumn{1}{l|}{x} & \multicolumn{1}{l|}{x} & \multicolumn{1}{l|}{x} & \multicolumn{1}{l|}{x} & \multicolumn{1}{l|}{x} & \multicolumn{1}{l|}{x} & \multicolumn{1}{l|}{x} & \multicolumn{1}{l|}{x} & \multicolumn{1}{l|}{x} & \multicolumn{1}{l|}{x} & x \\ \cline{3-19} 
\multicolumn{1}{|l|}{} & \multicolumn{1}{l|}{} & RD & \multicolumn{1}{l|}{x} & \multicolumn{1}{l|}{x} & \multicolumn{1}{l|}{x} & \multicolumn{1}{l|}{x} & \multicolumn{1}{l|}{x} & \multicolumn{1}{l|}{x} & \multicolumn{1}{l|}{x} & \multicolumn{1}{l|}{x} & \multicolumn{1}{l|}{x} & \multicolumn{1}{l|}{x} & \multicolumn{1}{l|}{x} & \multicolumn{1}{l|}{x} & \multicolumn{1}{l|}{x} & \multicolumn{1}{l|}{x} & \multicolumn{1}{l|}{x} & x \\ \cline{2-19} 
\multicolumn{1}{|l|}{} & \multicolumn{1}{l|}{\multirow{3}{*}{DN}} & BI & \multicolumn{1}{l|}{} & \multicolumn{1}{l|}{x} & \multicolumn{1}{l|}{x} & \multicolumn{1}{l|}{x} & \multicolumn{1}{l|}{} & \multicolumn{1}{l|}{x} & \multicolumn{1}{l|}{x} & \multicolumn{1}{l|}{x} & \multicolumn{1}{l|}{} & \multicolumn{1}{l|}{x} & \multicolumn{1}{l|}{x} & \multicolumn{1}{l|}{x} & \multicolumn{1}{l|}{} & \multicolumn{1}{l|}{} & \multicolumn{1}{l|}{} &  \\ \cline{3-19} 
\multicolumn{1}{|l|}{} & \multicolumn{1}{l|}{} & BIC & \multicolumn{1}{l|}{} & \multicolumn{1}{l|}{x} & \multicolumn{1}{l|}{x} & \multicolumn{1}{l|}{x} & \multicolumn{1}{l|}{} & \multicolumn{1}{l|}{x} & \multicolumn{1}{l|}{x} & \multicolumn{1}{l|}{x} & \multicolumn{1}{l|}{} & \multicolumn{1}{l|}{x} & \multicolumn{1}{l|}{x} & \multicolumn{1}{l|}{x} & \multicolumn{1}{l|}{} & \multicolumn{1}{l|}{} & \multicolumn{1}{l|}{} &  \\ \cline{3-19} 
\multicolumn{1}{|l|}{} & \multicolumn{1}{l|}{} & RD & \multicolumn{1}{l|}{} & \multicolumn{1}{l|}{x} & \multicolumn{1}{l|}{x} & \multicolumn{1}{l|}{x} & \multicolumn{1}{l|}{} & \multicolumn{1}{l|}{x} & \multicolumn{1}{l|}{x} & \multicolumn{1}{l|}{x} & \multicolumn{1}{l|}{} & \multicolumn{1}{l|}{x} & \multicolumn{1}{l|}{x} & \multicolumn{1}{l|}{x} & \multicolumn{1}{l|}{} & \multicolumn{1}{l|}{} & \multicolumn{1}{l|}{} &  \\ \cline{2-19} 
\multicolumn{1}{|l|}{} & \multicolumn{1}{l|}{\multirow{3}{*}{DS}} & BI & \multicolumn{1}{l|}{x} & \multicolumn{1}{l|}{x} & \multicolumn{1}{l|}{x} & \multicolumn{1}{l|}{x} & \multicolumn{1}{l|}{} & \multicolumn{1}{l|}{} & \multicolumn{1}{l|}{} & \multicolumn{1}{l|}{} & \multicolumn{1}{l|}{} & \multicolumn{1}{l|}{} & \multicolumn{1}{l|}{} & \multicolumn{1}{l|}{} & \multicolumn{1}{l|}{} & \multicolumn{1}{l|}{} & \multicolumn{1}{l|}{} &  \\ \cline{3-19} 
\multicolumn{1}{|l|}{} & \multicolumn{1}{l|}{} & BIC & \multicolumn{1}{l|}{x} & \multicolumn{1}{l|}{x} & \multicolumn{1}{l|}{x} & \multicolumn{1}{l|}{x} & \multicolumn{1}{l|}{} & \multicolumn{1}{l|}{} & \multicolumn{1}{l|}{} & \multicolumn{1}{l|}{} & \multicolumn{1}{l|}{} & \multicolumn{1}{l|}{} & \multicolumn{1}{l|}{} & \multicolumn{1}{l|}{} & \multicolumn{1}{l|}{} & \multicolumn{1}{l|}{} & \multicolumn{1}{l|}{} &  \\ \cline{3-19} 
\multicolumn{1}{|l|}{} & \multicolumn{1}{l|}{} & RD & \multicolumn{1}{l|}{x} & \multicolumn{1}{l|}{x} & \multicolumn{1}{l|}{x} & \multicolumn{1}{l|}{x} & \multicolumn{1}{l|}{} & \multicolumn{1}{l|}{} & \multicolumn{1}{l|}{} & \multicolumn{1}{l|}{} & \multicolumn{1}{l|}{} & \multicolumn{1}{l|}{} & \multicolumn{1}{l|}{} & \multicolumn{1}{l|}{} & \multicolumn{1}{l|}{} & \multicolumn{1}{l|}{} & \multicolumn{1}{l|}{} &  \\ \cline{2-19} 
\multicolumn{1}{|l|}{} & \multicolumn{1}{l|}{\multirow{3}{*}{DNS}} & BI & \multicolumn{1}{l|}{} & \multicolumn{1}{l|}{x} & \multicolumn{1}{l|}{x} & \multicolumn{1}{l|}{x} & \multicolumn{1}{l|}{} & \multicolumn{1}{l|}{} & \multicolumn{1}{l|}{} & \multicolumn{1}{l|}{} & \multicolumn{1}{l|}{} & \multicolumn{1}{l|}{} & \multicolumn{1}{l|}{} & \multicolumn{1}{l|}{} & \multicolumn{1}{l|}{} & \multicolumn{1}{l|}{} & \multicolumn{1}{l|}{} &  \\ \cline{3-19} 
\multicolumn{1}{|l|}{} & \multicolumn{1}{l|}{} & BIC & \multicolumn{1}{l|}{} & \multicolumn{1}{l|}{x} & \multicolumn{1}{l|}{x} & \multicolumn{1}{l|}{x} & \multicolumn{1}{l|}{} & \multicolumn{1}{l|}{} & \multicolumn{1}{l|}{} & \multicolumn{1}{l|}{} & \multicolumn{1}{l|}{} & \multicolumn{1}{l|}{} & \multicolumn{1}{l|}{} & \multicolumn{1}{l|}{} & \multicolumn{1}{l|}{} & \multicolumn{1}{l|}{} & \multicolumn{1}{l|}{} &  \\ \cline{3-19} 
\multicolumn{1}{|l|}{} & \multicolumn{1}{l|}{} & RD & \multicolumn{1}{l|}{} & \multicolumn{1}{l|}{x} & \multicolumn{1}{l|}{x} & \multicolumn{1}{l|}{x} & \multicolumn{1}{l|}{} & \multicolumn{1}{l|}{} & \multicolumn{1}{l|}{} & \multicolumn{1}{l|}{} & \multicolumn{1}{l|}{} & \multicolumn{1}{l|}{} & \multicolumn{1}{l|}{} & \multicolumn{1}{l|}{} & \multicolumn{1}{l|}{} & \multicolumn{1}{l|}{} & \multicolumn{1}{l|}{} &  \\ \hline
\multicolumn{3}{|l|}{IP 1D} & \multicolumn{1}{l|}{x} & \multicolumn{1}{l|}{x} & \multicolumn{1}{l|}{x} & \multicolumn{1}{l|}{x} & \multicolumn{1}{l|}{x} & \multicolumn{1}{l|}{x} & \multicolumn{1}{l|}{x} & \multicolumn{1}{l|}{x} & \multicolumn{1}{l|}{x} & \multicolumn{1}{l|}{x} & \multicolumn{1}{l|}{x} & \multicolumn{1}{l|}{x} & \multicolumn{1}{l|}{x} & \multicolumn{1}{l|}{x} & \multicolumn{1}{l|}{x} & x \\ \hline
\multicolumn{3}{|l|}{FP 2D} & \multicolumn{1}{l|}{x} & \multicolumn{1}{l|}{} & \multicolumn{1}{l|}{} & \multicolumn{1}{l|}{} & \multicolumn{1}{l|}{} & \multicolumn{1}{l|}{} & \multicolumn{1}{l|}{} & \multicolumn{1}{l|}{} & \multicolumn{1}{l|}{} & \multicolumn{1}{l|}{} & \multicolumn{1}{l|}{} & \multicolumn{1}{l|}{} & \multicolumn{1}{l|}{} & \multicolumn{1}{l|}{} & \multicolumn{1}{l|}{} &  \\ \hline
\end{tabular}%
}
\label{tab1}
\end{table}
 
\subsection{Architecture and training of the neural network}\label{impl}

For the purpose of validity, the neural network architecture is equal for all tests and has 5 hidden layers, 64 neurons per layer. The open-source Python library NeuralUQ is used to apply UQ methods to our models \cite{neuraluq}. The HMC is used for posterior sampling, the leapfrog step is L = $50\delta t$, the initial time step is $\delta t = 0.01$, the burn-in steps are set to 1000 and a total of 1500 samples are taken. The only variable parameter between the different tests is the number of collocation points and input conditions used to solve different PDEs, which are specified in Table \ref{tab:numpoint}. In order to illustrate the results, we show the snapshots at time $t=0.5$ for each of the 1D cases presented further. 

\begin{table}[ht]
    \centering
    \caption{Numbers of training and collocation points used for different PDEs in different computational experiments}
\resizebox{\textwidth}{!}{  
    \begin{tabular}{|l|c|c|c|c|c|c|}
    \hline
    \parbox[t]{5mm}&{{\rotatebox[origin=c]{90}{Boundary Conditions (BC)}}} 
    & \parbox[t]{2mm}{{\rotatebox[origin=c]{90}{Initial Conditions (IC)}}}  & \rotatebox[origin=c]{90}{\parbox[t]{3cm}{Collocation~points at~the~interface (CDC)}} & \parbox[t]{2mm}{{\rotatebox[origin=c]{90}{Collocation points}}} &\parbox[t]{2mm}{{\rotatebox[origin=c]{90}{Data points}}} &
   \rotatebox[origin=c]{90}{\parbox[t]{3cm}{Data points at the interface}}\\ \hline    
 Allen-Cahn BPINN & 40 & 80 & - & 60 & 60 & - \\ \hline
  Allen-Cahn \$PINN  & 20/20 & 40/40 & 20 & 30/30 & 30/30& 20\\ \hline

Allen-Cahn BPINN, IP  & - & - & - & 300 & 300 & - \\ \hline
  Allen-Cahn \$PINN, IP  & - & - & 40 & 150/150 & 150/150 & - \\ \hline
 Burger's BPINN & 40 & 150 & - & 300 & 250 & - \\ \hline
 Burger's \$PINN  & 20/20 & 40/40 & 20 & 100/100 & 30/30 & 20 \\ \hline
  Fokker-Planck BPINN & 40 & 80 & - & 60 & 60 & - \\ \hline
   Fokker-Planck \$PINN & 20/20 & 40/40 & 20 & 30/30 & 30/30 & 20\\ \hline
   Fokker-Planck \$PINN, DS   & 20/20  & 60/20  & 20 &  40/20 &40/20 &20\\ \hline
   Fokker-Planck \$PINN, DNS & 20/20  & 60/20  & 20 &  40/20 &40/20 &20 \\ \hline
    Fokker-Planck BPINN, IP & - & - & - & 300 & 300 & - \\ \hline
   Fokker-Planck \$PINN, IP  & - & - & 40 & 150/150 & 150/150 & -  \\ \hline
    Fisher KPP BPINN & 40 & 80 & - & 60 & 60 & -\\ \hline
    Fisher KPP \$PINN & 20/20 & 40/40 & 20 & 30/30 & 30/30 & 20\\ \hline 
    \end{tabular}
    \label{tab:numpoint}}
\end{table}

\section{Results} \label{results}
This section presents the results for selected computational experiments that highlight the implementation of \$PINN, provided parameters outlined in Section \ref{sec:computational_experiments}. The forward problem solution with \$PINN and BPINN are compared given variable levels of added noise using Allen-Cahn equation. 
The results for variable noise levels, different PDEs and variable noise levels are presented, followed by results with variable domain sizes. The inverse problem and the 2D forward problem solutions are shown at the end. 

\subsection{Added noise - DN}
To show the results of \$PINN performance at different levels of added noise, we use the Allen-Cahn equation. At first, we show the performance of \$PINN with different levels of added noise using two training strategies described in Subsection \ref{impl}. In this case, the noise levels are kept equal across the solution space. In the second case, we also introduce the variability in noise between the domains, where it is added only to one of the analyzed subdomains or at different levels.


\subsubsection{Variable levels of added noise}\label{VLofnoise}
\begin{figure}[H]
    \centering
    \includegraphics[width=1.0\linewidth]{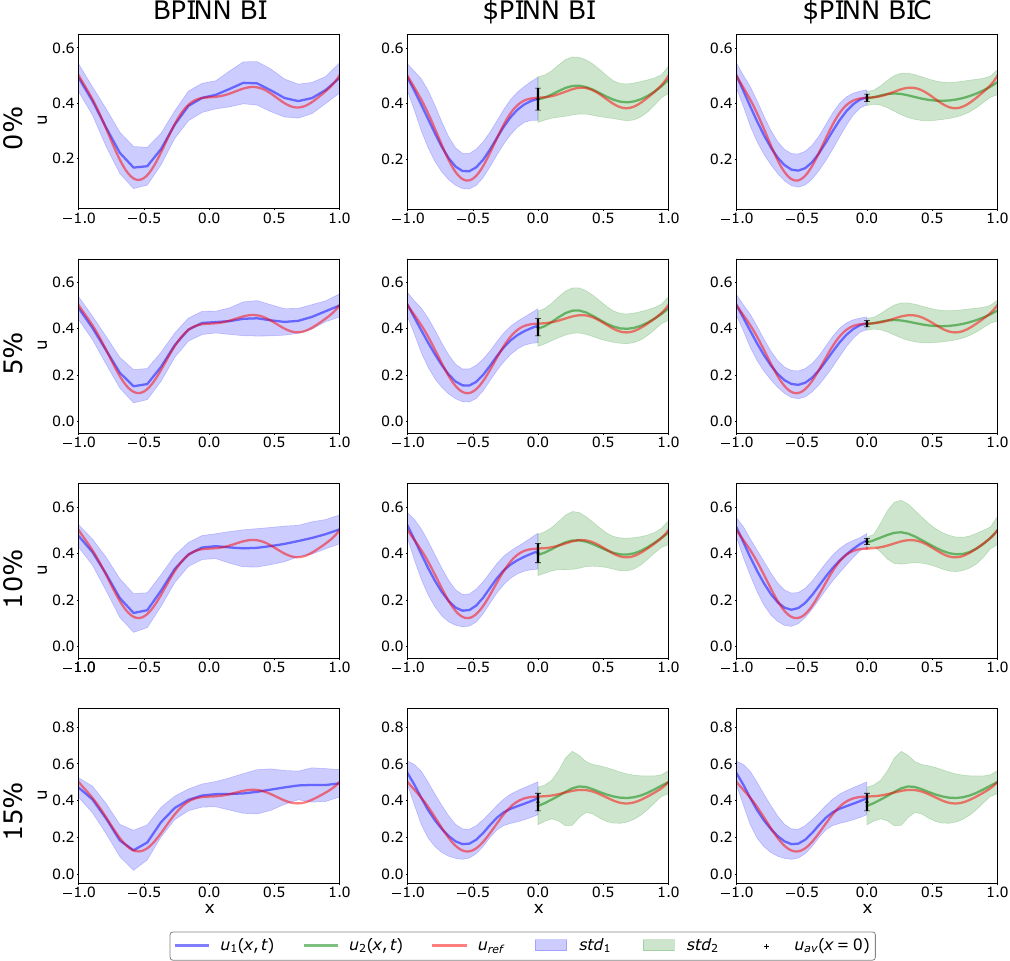}
    \caption{Allen-Cahn with BPINN and \$PINN using IC, BC and CDC.}
    \label{fig:AC_bc_ic_all}
\end{figure}

Figure \ref{fig:AC_bc_ic_all} shows the comparison of the BPINN BI case results (left column) with the corresponding \$PINN BI results (middle column) and BIC results (right column). BPINN exhibits robustness in performance throughout the four noisy scenarios. \$PINN also shows accurate predictions, closely matching the reference solution even with higher noise levels. Moreover, adding more points along the domain interface in the BIC sub-case results in further improvement by noticeably reducing the epistemic uncertainty.



\begin{figure}[H]
    \centering
    \includegraphics[width=0.67\linewidth]{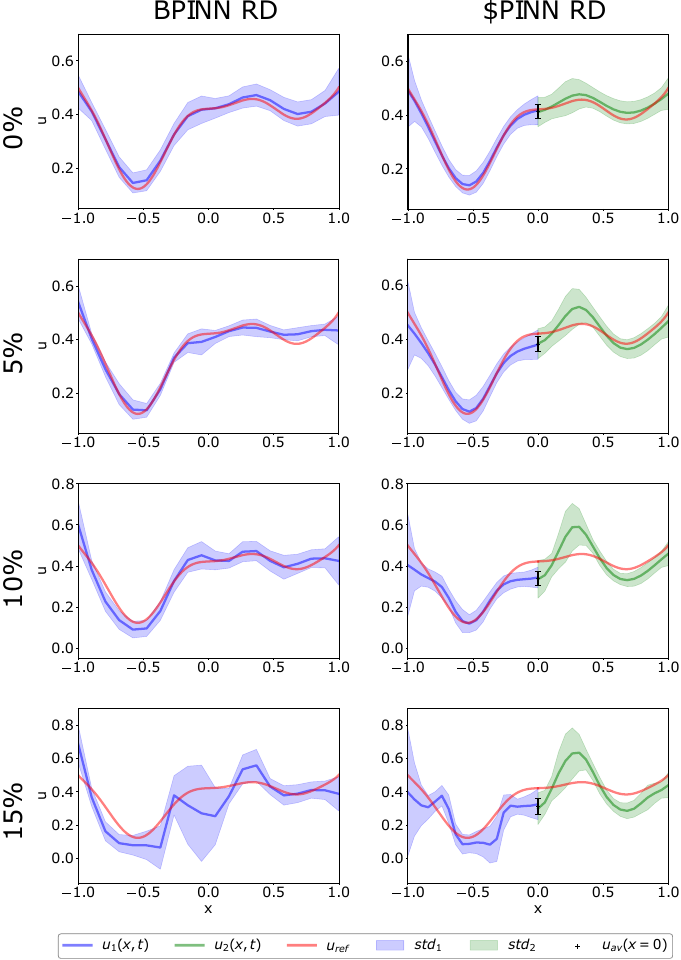}
    \caption{Allen-Cahn with BPINN and \$PINN using data points.}
    \label{fig:AC_rd_all}
\end{figure}

Figure \ref{fig:AC_rd_all} compares the BPINN RD case with the \$PINN. Both models capture the solution well for lower noise levels, especially BPINN. When reaching 15\% noise, the models struggle to fit the solution. 
The presented cases show that domain decomposition within the Bayesian framework can be effectively employed without compromising accuracy.

\subsubsection{Different levels of added noise between the domains}
\begin{figure}[H]
    \centering
    \includegraphics[width=1.0\linewidth]{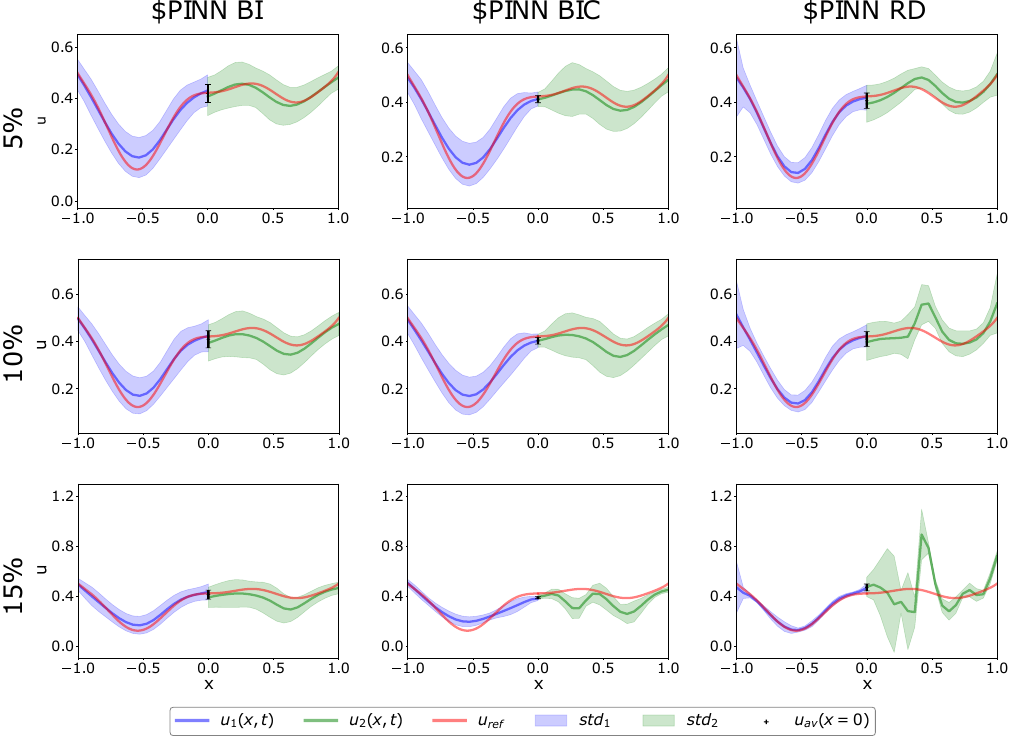}
    \caption{Allen-Cahn with \$PINN used for the three cases (BI, BIC, RD) with variable noise levels in the second domain.}
    \label{fig:AC_dn_all}
\end{figure}

Figure \ref{fig:AC_dn_all} presents the results for testing \$PINN with changed levels of added noise in the domains. In Figure \ref{fig:AC_dn_all} the left side of each plot has 0\% of added noise, while the noise level on the right side is variable. 

The model maintains good performance for noise levels of 5\% and 10\% within the noisy domain. However, as the noise increases to 15\%, performance declines, similar to the results in Subsection \ref{VLofnoise}, where both domains were subjected to the same noise level. 
For the case where only data points are used, in the right column, performance for the second domain decreases from noise levels of 10\%, reaching more random behavior at 15\%. Compared to the results where both domains shared the same noise levels, accuracy was improved for lower noise levels.
Therefore, incorporating data of varying quality can help the overall predictions since localized noise in one domain does not necessarily compromise the accuracy of the noise-free domain.


\subsection{Test on different PDEs}\label{tondifPDE}
The performance of \$PINN is tested on three additional PDEs also with the variable noise levels as in Subsection \ref{VLofnoise}.


\subsubsection{Burgers' equation}
\begin{figure}[H]
    \centering
    \includegraphics[width=1.0\linewidth]{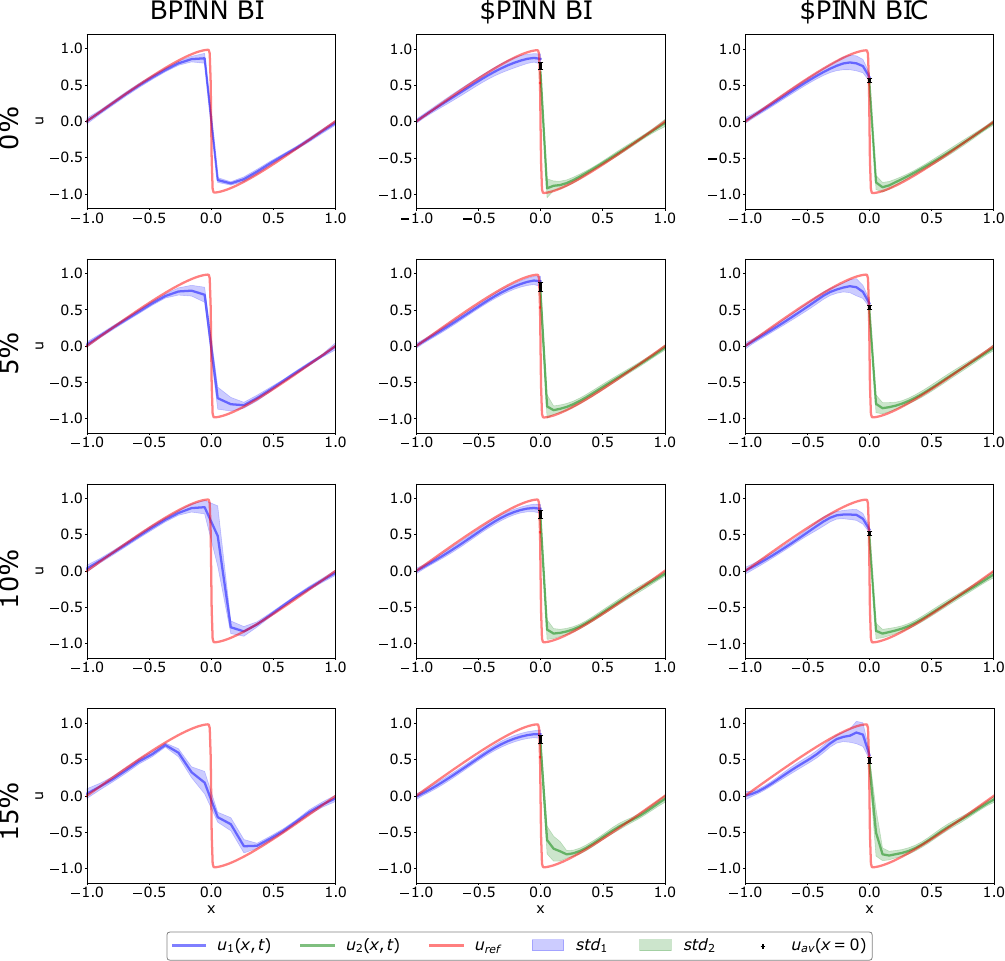}
    \caption{Burgers for BPINN and \$PINN for IC, BC and CD.}
    \label{fig:BR_bc_ic_all}
\end{figure}

\begin{figure}[H]
    \centering
    \includegraphics[width=0.67\linewidth]{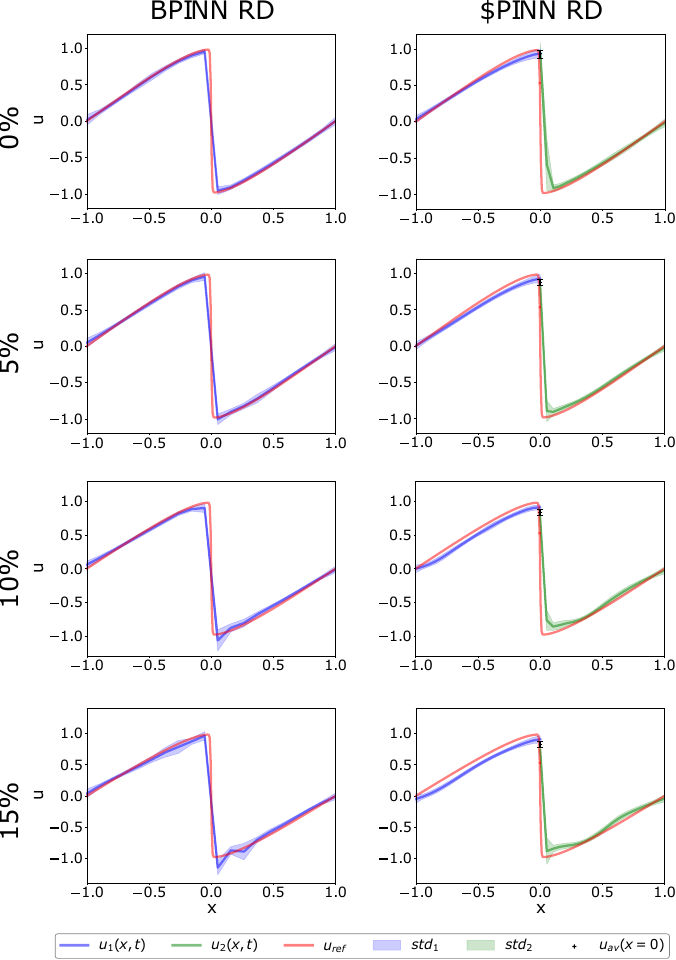}
    \caption{Burgers for BPINN and \$PINN using data points.}
    \label{fig:BurgersBpinnvsPinn}
\end{figure}

The case with Burgers' implementation is the only one, where we use different numbers of collocation points, initial conditions, and data points in the RD case, the exact numbers are specified in Table \ref{tab:numpoint}. This big difference is caused by the training of BPINN and its inability to approximate the function well with lower numbers. However, as shown in
Figures~\ref{fig:BR_bc_ic_all}~and~\ref{fig:BurgersBpinnvsPinn}, \$PINN 
gives a good approximation for Burgers' using fewer points. As shown in the right column of Figure \ref{fig:BR_bc_ic_all}, incorporating data points in \$PINN from the boundaries, the initial condition, and the shared domain leads to a slight increase in overfitting, but effectively reduces uncertainty near the interface. BPINN is more sensitive to higher noise levels (e.g. 15\%), with the increasing variance and deviation, while \$PINN maintains better robustness, with lower uncertainty and closer alignment with the reference. Figure \ref{fig:BurgersBpinnvsPinn} demonstrates the benefit of distributing random data points throughout the domain, and both models work better approximation compared to the previous case.  


\subsubsection{Fokker-Planck equation}
\begin{figure}[H]
    \centering
    \includegraphics[width=1.0\linewidth]{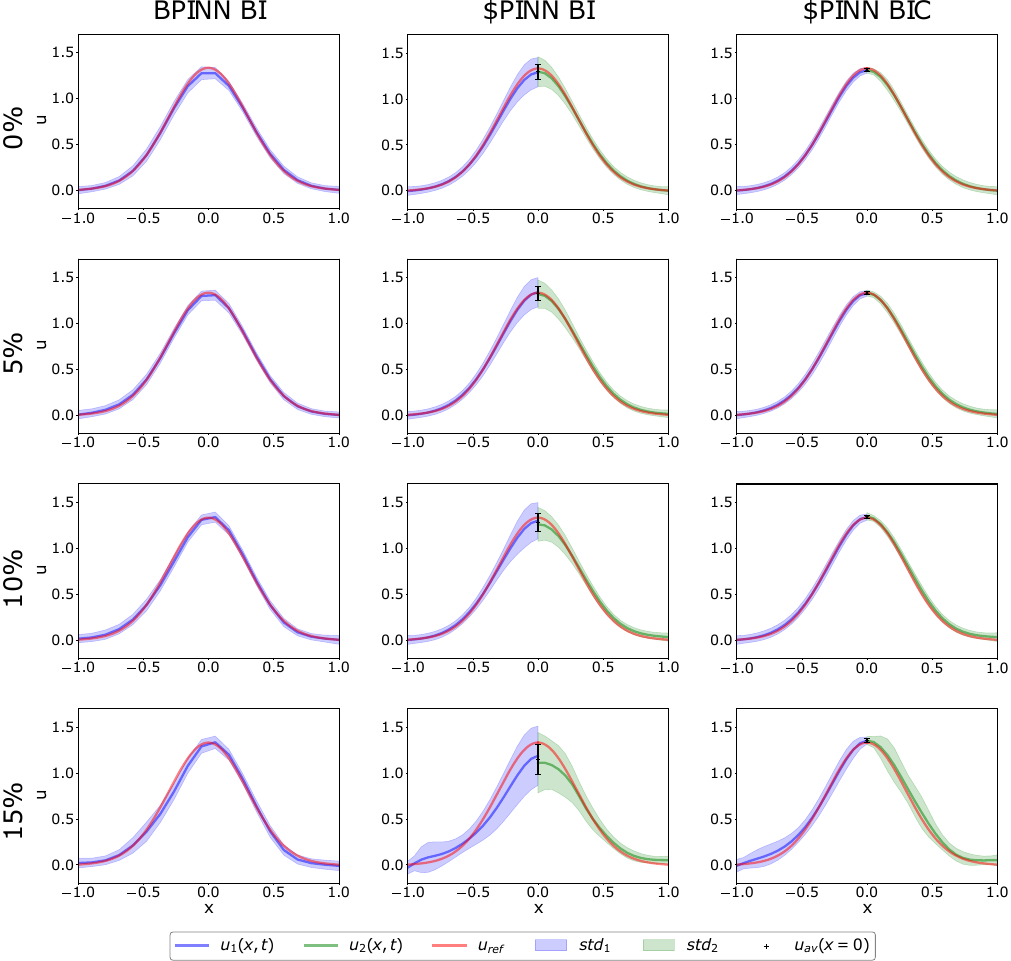}
    \caption{Fokker-Planck for BPINN and \$PINN for IC, BC and CD.}
    \label{fig:FP_bc_ic_all}
\end{figure}

\begin{figure}[H]
    \centering
    \includegraphics[width=0.67\linewidth]{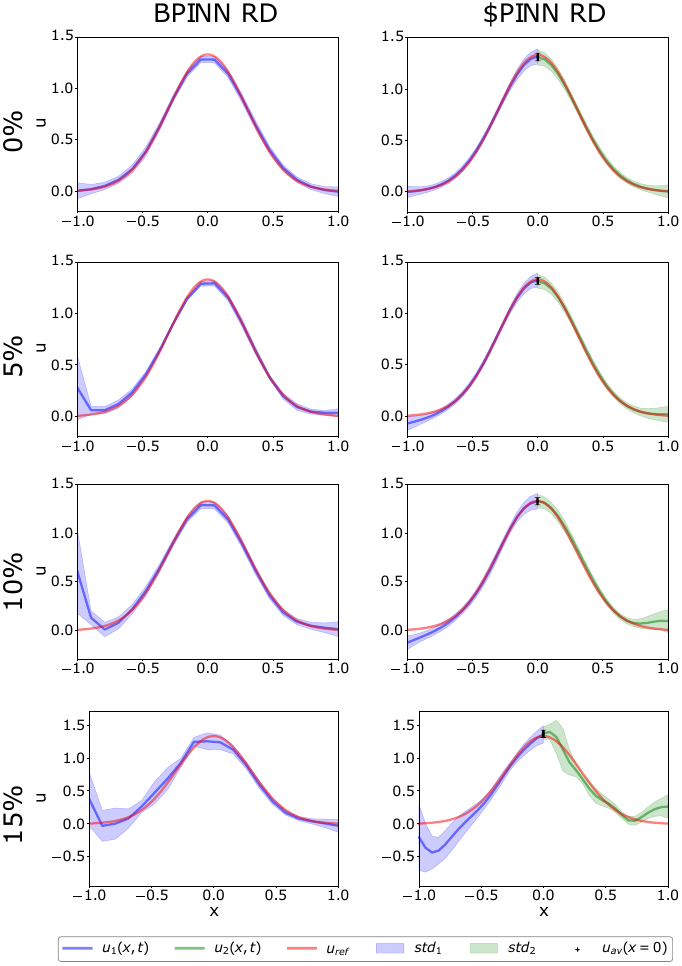}
    \caption{Fokker-Planck for BPINN and \$PINN using data points.}
    \label{fig:FP_rd_all}
\end{figure}

A comparison of BPINN and \$PINN performance for the Fokker-Planck equation is presented in Figures \ref{fig:FP_bc_ic_all} and \ref{fig:FP_rd_all}. As shown in Figure \ref{fig:FP_bc_ic_all}, including data points at the interface in \$PINN again shows higher improvement. It significantly reduces epistemic uncertainty, which propagates across both subdomains. In this way, \$PINN matches the reference solution closely, achieving results comparable to BPINN. As noise increases, \$PINN BI progressively worsens its performance, showing a more significant effect of random uncertainty when 15\% noise is reached. \$PINN BIC also slightly suffers from the highest noise level. However, its predictions are comparable to those of the BPINN case. 
When only randomly distributed data points are used, as in Figure \ref{fig:FP_rd_all}, both models become more prone to overfitting, predicting less accurate results with increasing noise levels. BPINN generally exhibits greater difficulty capturing the solution behavior near the domain boundaries compared to \$PINN. However, at 15\% noise, both methods show a noticeable drop in quality, and no substantial qualitative advantage is observed for either approach.

\subsubsection{Fisher-KPP equation}
\begin{figure}[H]
    \centering
    \includegraphics[width=1.0\linewidth]{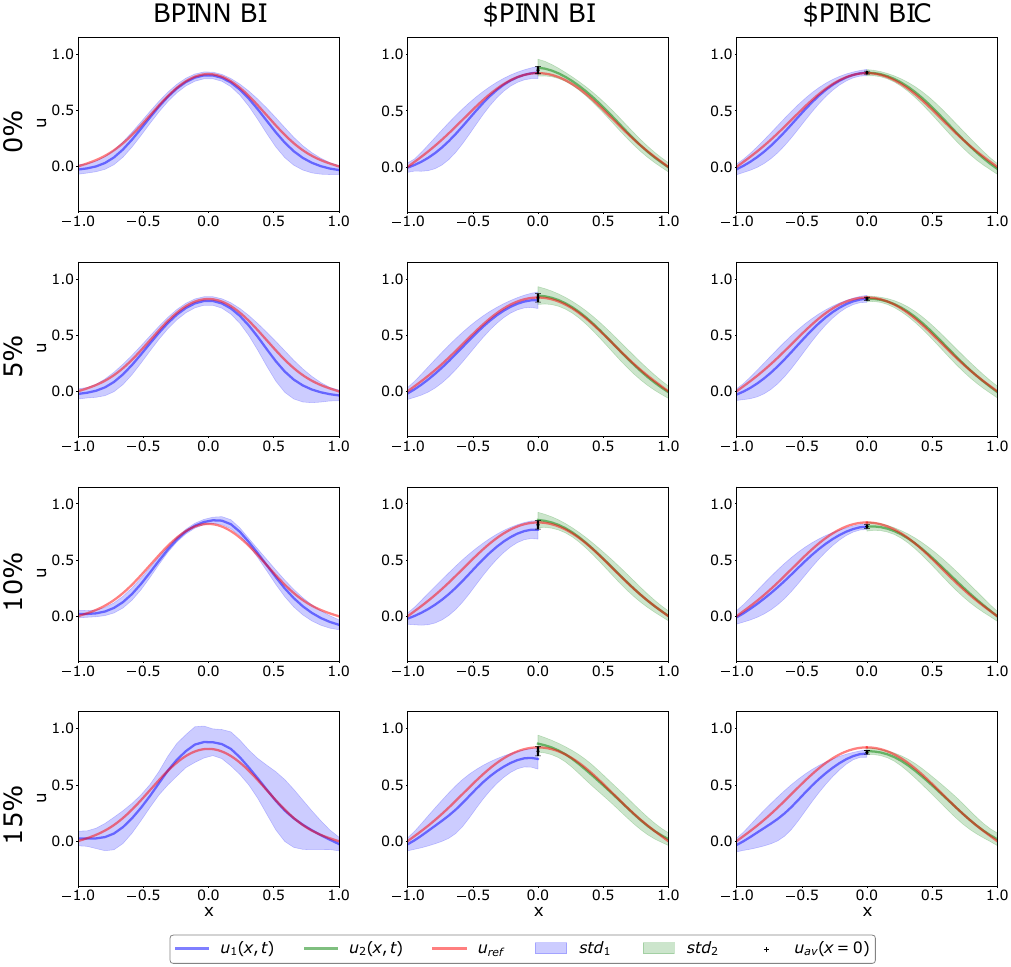}
    \caption{Fisher-KPP for BPINN and \$PINN for IC, BC and CD.}
    \label{fig:FKPP_bc_ic_all}
\end{figure}

\begin{figure}[H]
    \centering
    \includegraphics[width=0.67\linewidth]{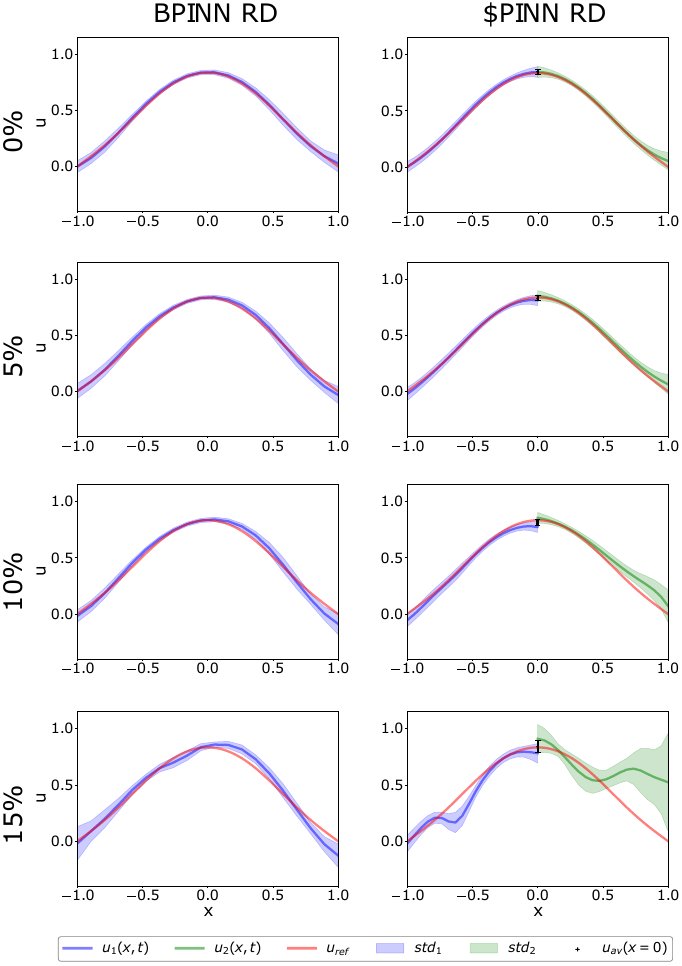}
    \caption{Fisher-KPP for BPINN and \$PINN using data points.}
    \label{fig:FKPP_rd_all}
\end{figure}

The results of \$PINN application and the comparison with BPINN are shown in Figures \ref{fig:FKPP_bc_ic_all} and \ref{fig:FKPP_rd_all} for both training scenarios. 
We observe a notable reduction in epistemic uncertainty after additional data points are added at the interface for \$PINN as shown in the right column in Figure \ref{fig:FKPP_bc_ic_all}.
In contrast, when using only random data points without specifying the initial condition, boundaries, or interface, an increased epistemic uncertainty appears at the left boundary of the first subdomain and the right boundary of the second subdomain as shown in Figure \ref{fig:FKPP_rd_all}. In this case, BPINN behaves more or less similarly to \$PINN. However, overall, \$PINN outperforms BPINN by exhibiting lower epistemic uncertainty, as highlighted by the horizontal comparison, and reduced aleatoric uncertainty as noise levels increase.


\subsection{Variable domain and noise levels test}
To eliminate further biases we use Fokker-Planck equation to illustrate the performance of \$PINN for variable domain sizes 
First, we keep equal noise value across the domains and later introduce noise variability between domains, adding noise to only one subdomain. 




\subsubsection{Different domain size - DS}
\begin{figure}[H]
    \centering
    \includegraphics[width=1.0\linewidth]{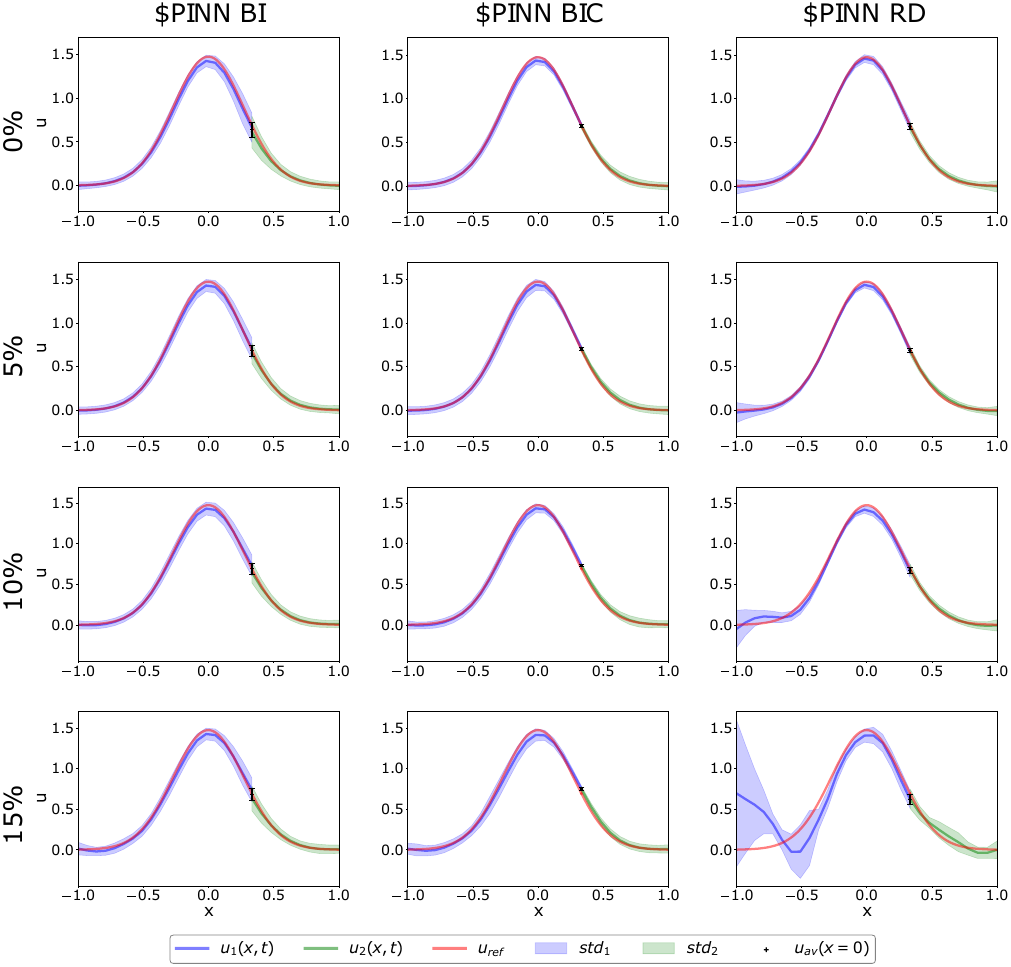}
    \caption{Fokker-Planck with \$PINN used for the three cases (BI, BIC, RD) with different domain sizes (DS).}
    \label{fig:FP_ds_all}
\end{figure}

The results for \$PINN BI, BIC, and RD for different subdomain sizes for the Fokker-Planck equation are given in Figure \ref{fig:FP_ds_all}. In this case, subdomain 1 occupies two-thirds of the entire space solution. The noise is spread equally across the two subdomains. When only initial and boundary conditions are provided, in the left column, the model fit is fairly good and has rather low standard deviation. 
The accuracy is further improved near the interface, when additional points are added to the \$PINN BIC, displayed in the middle column of Figure \ref{fig:FP_ds_all}. Finally, when using data points as input, the accuracy drops at higher added noise levels, specifically 10\% and 15\%. Therefore, we can assume that disproportional domains do not affect \$PINN performance, and in certain scenarios may even have some advantages compared to the equal division.

\begin{figure}[H]
    \centering
    \includegraphics[width=1\linewidth]{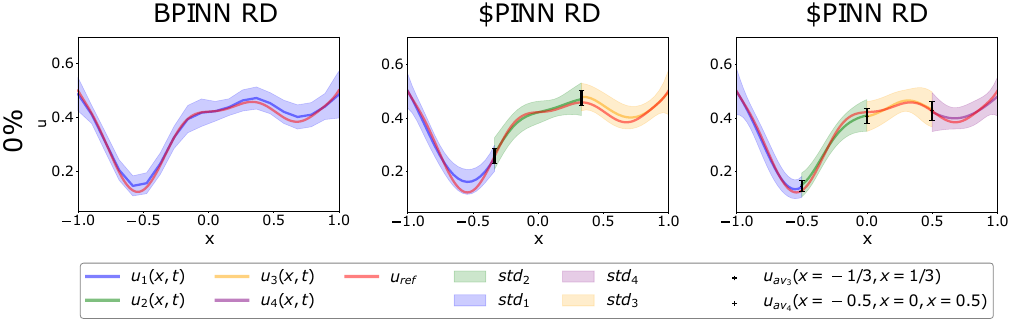}
    \caption{Allen-Cahn for BPINN and \$PINN with 3 and 4 subdomains using data points.}
    \label{fig:AC_34_all}
\end{figure}

We have also performed a test with 3 domains and 4 domains, the results for the Allen-Cahn equation are shown in Figure \ref{fig:AC_34_all}. In both cases the uncertainty level is slightly higher at the interface, however, it is likely that the fit can be improved by adding more collocation and data points at the interface. The standard deviation is acceptable and is not much different from BPINN or case with 2 subdomains. However, it should to be noted that the problem size in this test was fairly small and for larger problems setups with larger numbers of subdomains may result in improved performance due to the availability of sufficient computational resources.  

\subsubsection{Different noise levels in variable domains}

\begin{figure}[H]
    \centering
    \includegraphics[width=1.0\linewidth]{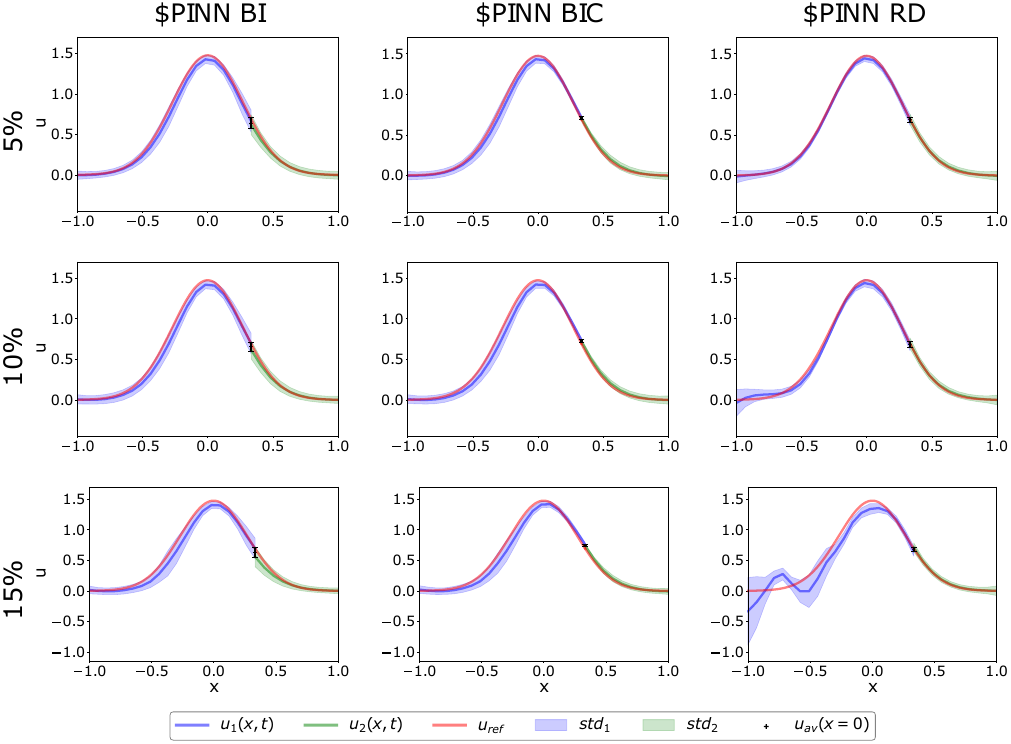}
    \caption{Fokker-Planck with \$PINN used for the three cases (BI, BIC, RD) for a combination of different noise levels and domain sizes (DNS).}
    \label{fig:FP_ds_dn_all}
\end{figure}

Figure \ref{fig:FP_ds_dn_all} presents the results obtained when combining varying noise levels with subdomains of different sizes for the Fokker-Planck equation. In this case, different noise levels (5\%, 10\%, and 15\%) are added to the larger subdomain, while the smaller domain is kept noise-free. The left column shows the results obtained for \$PINN BI, where performance remains accurate. The performance is more robust, when introducing points at the interface, as in the case of \$PINN BIC, reducing the standard deviation. Finally, in the right column, where only data points in the overall solution space are used, the prediction accuracy decreases as more noise is added in the first subdomain. In particular, the left boundary presents the worst performance with a higher standard deviation.  Comparing the results obtained in Figure \ref{fig:FP_ds_all} and \ref{fig:FP_ds_dn_all}, we can conclude that having noise localized on the larger side of the domain only leads to more accurate predictions, especially if only data points are available in the domain.

\subsection{Inverse problem}\label{secinverse}
To show the performance of \$PINN applied to the inverse problems, we infer the diffusion coefficient $D$ using three techniques. First, we present the results where we estimate the unknown parameters for each subdomain, as expressed in Equations \eqref{inverse1}-\eqref{eq:spinn_inv2}. We then improve the results by adding soft and hard constraints, as defined in Equations \eqref{eq:soft_cond}-\eqref{eq:spinn_inv2_hc}. 
\subsubsection{Allen-Cahn equation}
\begin{figure}[H]
     \centering
     \begin{subfigure}[b]{0.49\textwidth}
         \centering
         \includegraphics[width=\textwidth]{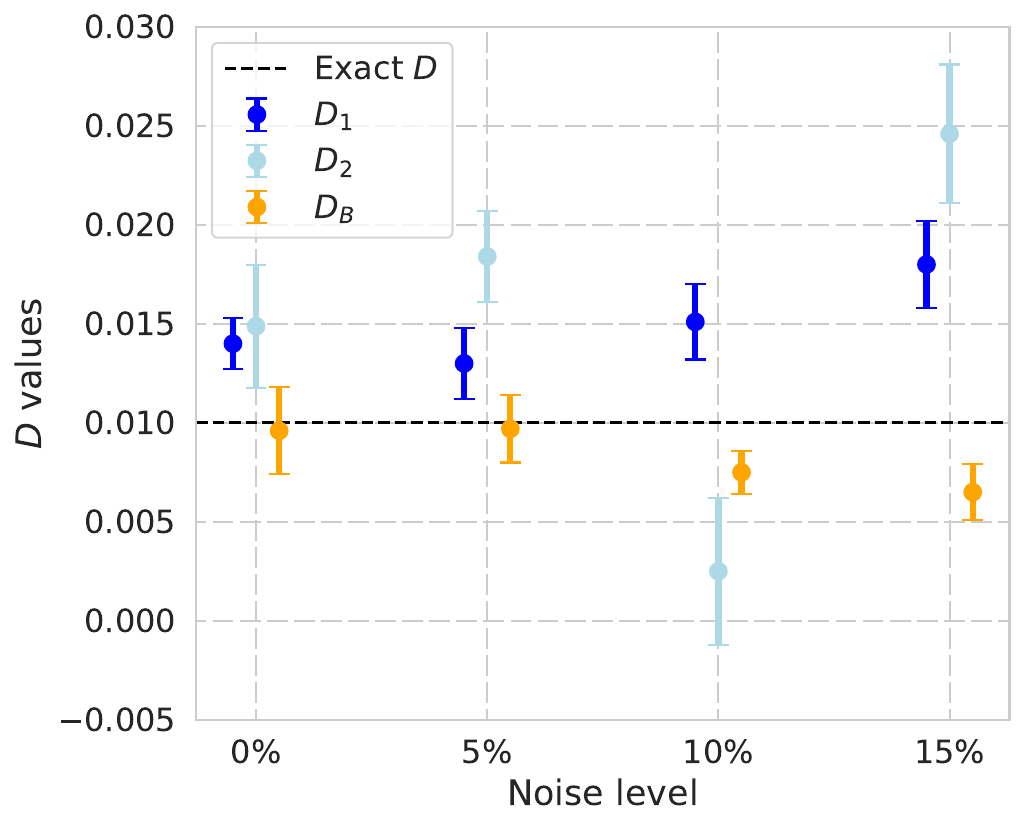}
         \caption{Without constraints}
         \label{fig:coeffD}
     \end{subfigure}
     \begin{subfigure}[b]{0.49\textwidth}
         \centering
         \includegraphics[width=\textwidth]{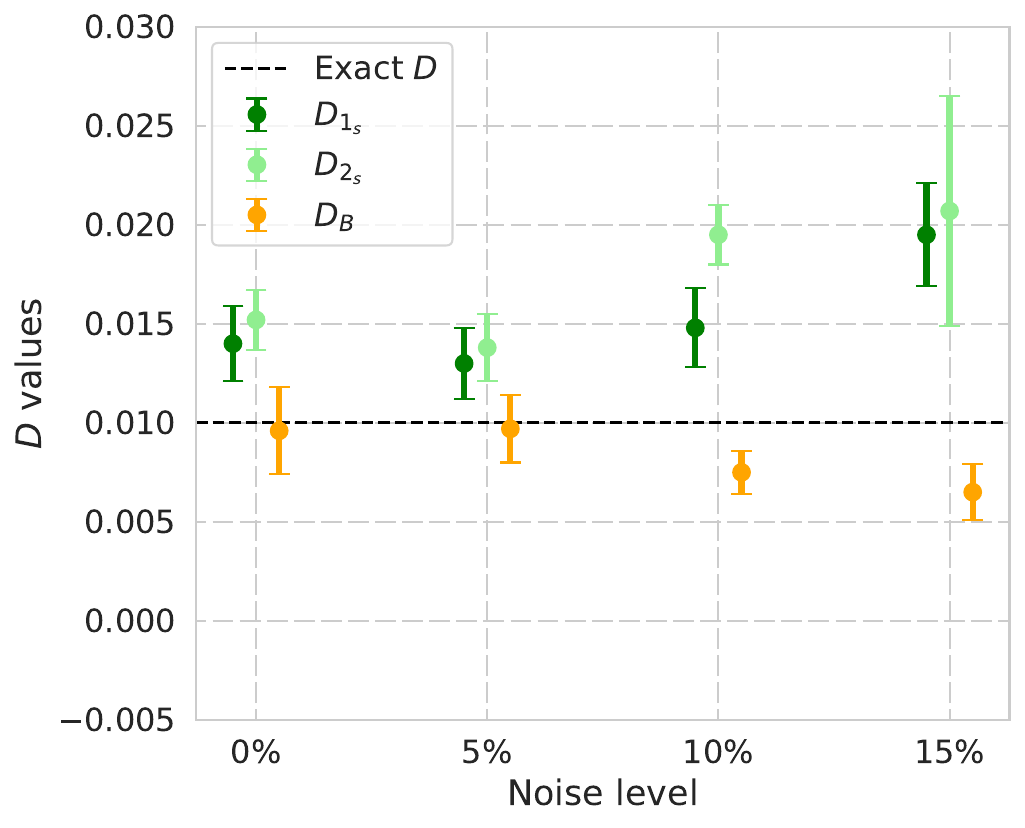}
         \caption{Soft constraints}
         \label{fig:coeffDsoft}
     \end{subfigure}
     \begin{subfigure}[b]{0.49\textwidth}
         \centering
         \includegraphics[width=\textwidth]{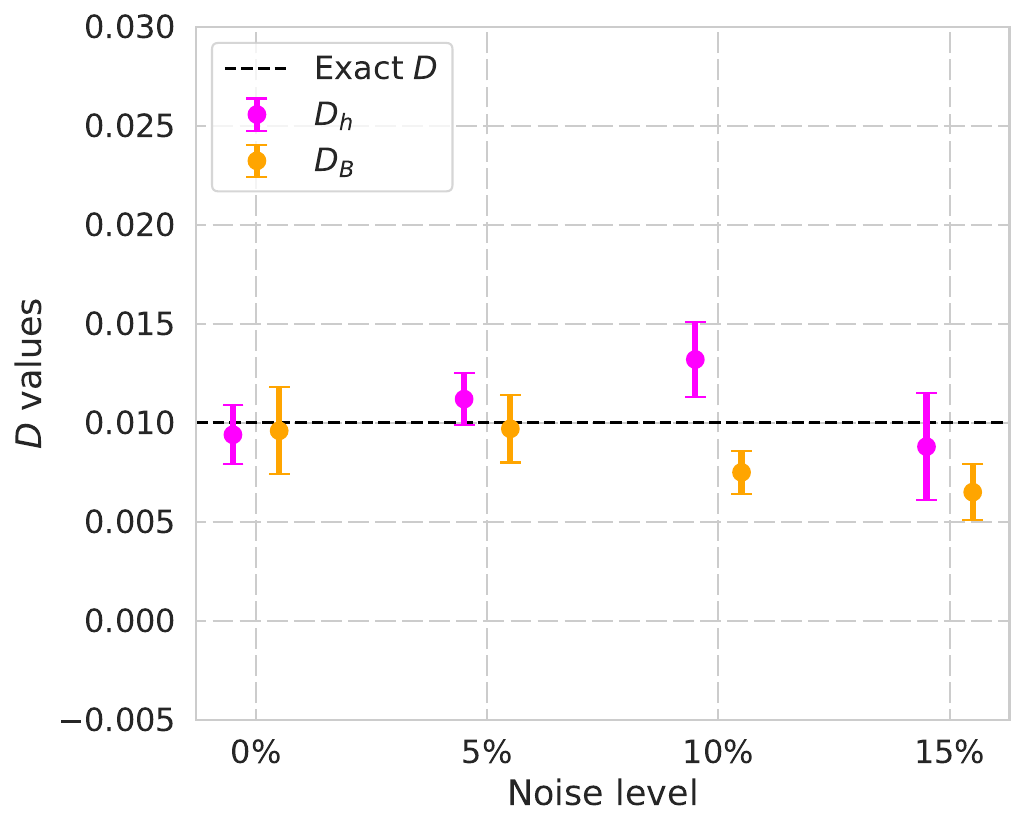}
         \caption{Hard constraints}
         \label{fig:coeffhard}
     \end{subfigure}
     \caption{Recovery of the diffusion coefficient ($D = 0.01$) (a) without specific constraints (b) with soft and (c) with hard constraints as a function of the noise level.}
    \label{fig:coeffD_all}
\end{figure}

\begin{figure}[H]
    \centering
    \includegraphics[width=0.95\linewidth]{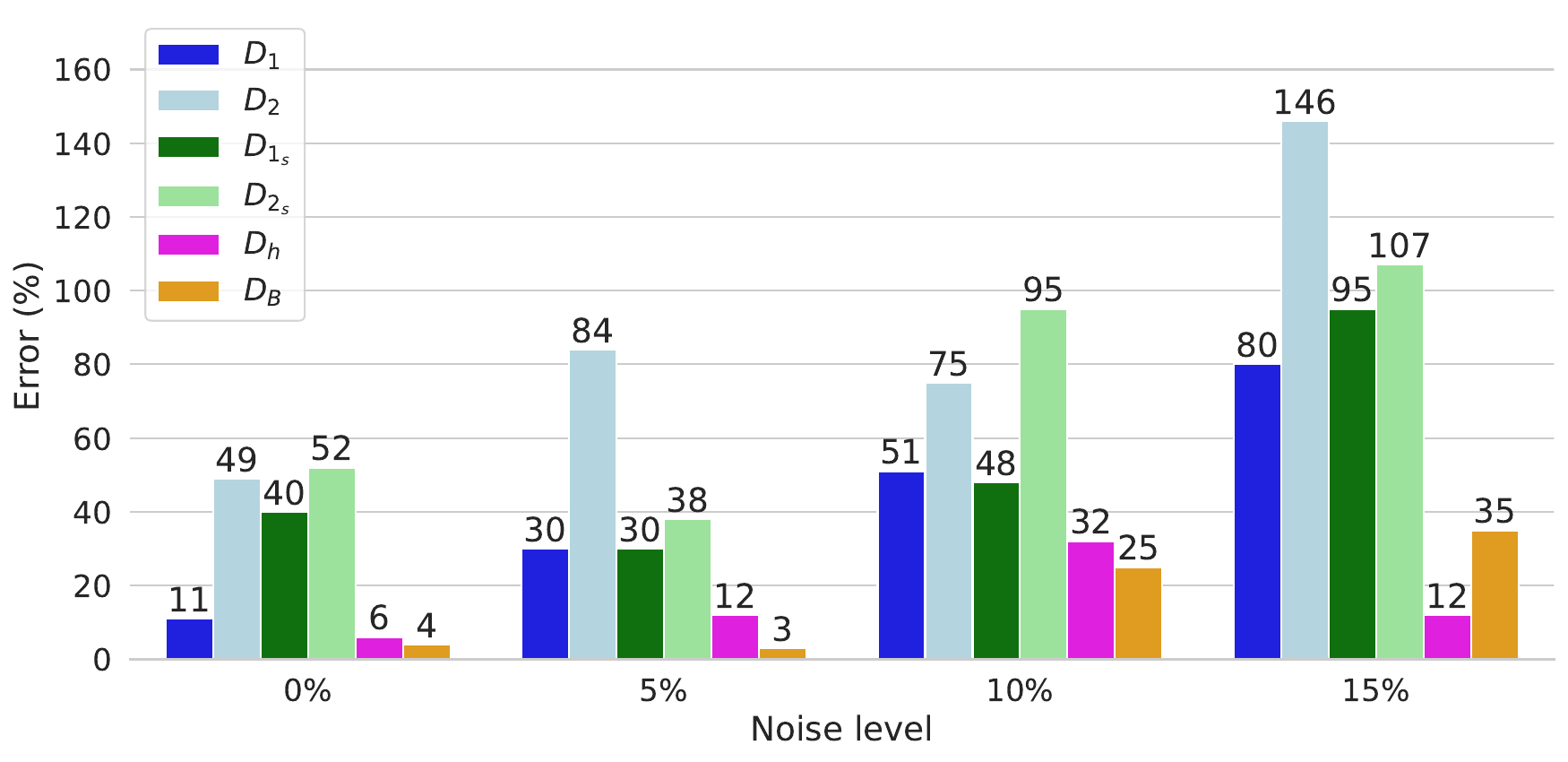}
    \caption{Absolute error (\%) of the diffusion coefficient recovery for the Allen-Cahn equation for \$PINN without constraints (dark and light blue), with soft constraint (dark and light green), with hard constraint (pink), and BPINN (orange).}
    \label{fig:AC_inverse_D}
\end{figure}

Figure \ref{fig:coeffD_all} shows the prediction errors for the three cases for \$PINN and the one for BPINN: Figure \ref{fig:coeffD} shows the case without constraints, where the coefficients $D_1$ and $D_2$ stand for predictions in each subdomain and $D_B$ is the estimation of $D$ using BPINN. The predicted diffusion coefficients $D_1$ and $D_2$ are different, when they should be closer in value. Furthermore, the difference between $D_1$, $D_2$ and the actual $D=0.01$ (solid black line) increases with noise. BPINN performs better than \$PINN. 
When soft constraints are added in Figure \ref{fig:coeffDsoft}, the diffusion coefficients, defined as $D_{1_s}$ and $D_{2_s}$ and represented in dark and light green, respectively, match their reciprocal values more closely. The diffusion coefficients of the two subdomains are now forced to be equal at the interface, but it does not give sufficient improvements. Therefore, \$PINN estimation of $D_h$ (pink) is improved by imposing the hard constraint as seen in Figure \ref{fig:coeffhard}.
Figure \ref{fig:AC_inverse_D} shows the Absolute error (\%) of the inferred $D$ for \$PINN without constraints (dark and light blue), with soft constraints (dark and light green), with hard constraint (pink), and BPINN (orange) compared to the actual value for the four noise levels. Imposing hard constraints results in significant improvement for \$PINN. 

\subsubsection{Fokker-Planck}
\begin{figure}[H]
    \centering
    \includegraphics[width=0.45\linewidth]{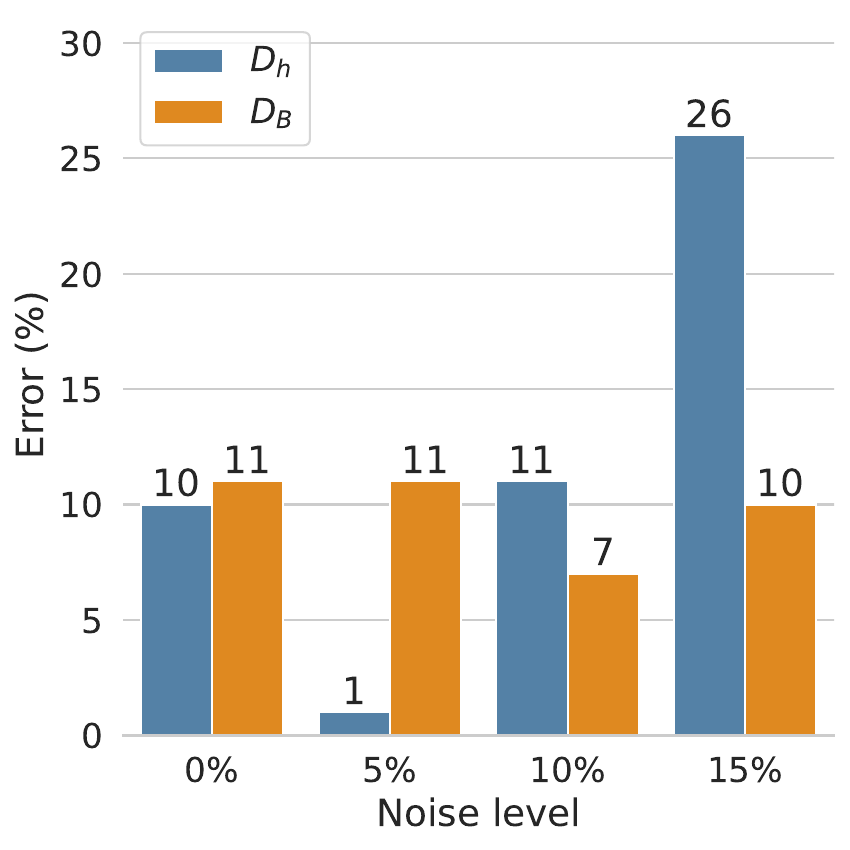}
    \caption{Absolute error (\%) of the diffusion coefficient recovery for the Fokker-Planck equation for \$PINN with hard constraint (blue) and BPINN (orange) for variable noise.}
    \label{fig:FP_inverse_D}
\end{figure}

The best performing inverse \$PINN model using hard constraints is now tested using Fokker-Planck PDE. 
Figure \ref{fig:FP_inverse_D}  compares BPINN and \$PINN using absolute error and variable noise, where $D_h$ (blue) for \$PINN and  $D_B$ (orange) for BPINN. 
\$PINN is slightly worse than BPINN on average, except for noise equal to 15 \%, where its performance drops significantly. 

\subsection{2D problem}
\begin{figure}[H]
    \centering
    \includegraphics[width=1\linewidth]{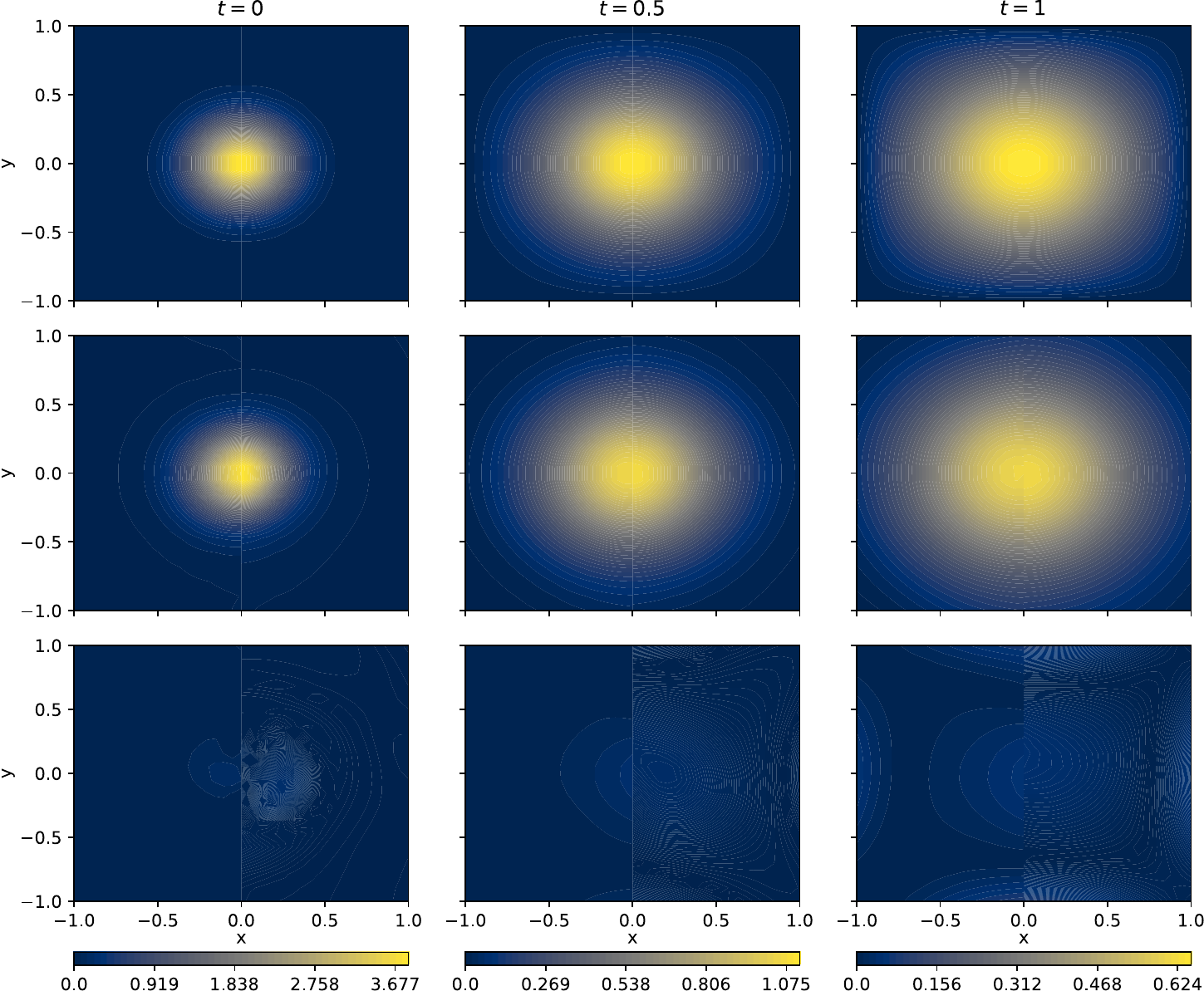}
    \caption{Fokker-Planck for BPINN and \$PINN 2D forward problem using data points.}
    \label{fig:FP_2D}
\end{figure}

Extending the implementation of \$PINN to 2D spatial domains with time is more complex and realistic testbed. 
Figure \ref{fig:FP_2D} shows validation of the method in a 2D space with time, $(x,y,t)$, setting for the Fokker-Planck equation. 
Given 300 residual points randomly distributed in space and time, and 60 initial points randomly distributed in space, we found correct performance of the forward problem without noise. There is a low standard deviation in all the solution, being a little higher with a bigger discrepancy between subdomain at the interface and in the boundaries of the domains. However, this test confirms that using \$PINN in 2D is feasible. Using a GPU and extra data points would allow even better accuracy. 

\section{Conclusions}
This paper proposes a novel method, \$PINN, incorporating Bayesian inference with domain decomposition. \$PINN exploits the robustness of BPINN to handle both epistemic and aleatoric uncertainties and the scalability of cPINN to large-scale and complex problems. We test our model on both forward and inverse problems on nonlinear PDEs, including the Allen-Cahn equation, the Burgers' equation, the Fokker-Planck equation, and the Fisher-KPP equation. We show different training strategies for all these cases: using initial and boundary conditions with collocation points across the subdomains (BI), and using randomly sampled data points both from the solution space and at the interface along with collocation points (RD). A sub-case is also introduced for the BI case, adding extra points at the interface (BIC). Several tests are examined with variable noise levels and domain sizes. All the tests prove that the proposed method does not lose accuracy compared to the BPINN model. In particular, adding extra points at the interface results in a more robust approximation in most cases. In the case of the Burgers' equation, where the domain interface is placed at the shock, we also observe an improvement in the performance of \$PINN over the BPINN predictions. For the inverse problem, we have tested three approaches to estimate the unknown parameter. First, we set up an unknown parameter for each subdomain without using any conditions. As this approach yields unbalanced predictions of the coefficients, extra conditions are introduced. A soft constraint is tested to impose more equality in the two values, and a hard constraint is tested, where only one unknown parameter is defined for both subdomains. The results enforcing hard constraints perform better than with soft constraints and give predictions comparable to the BPINN model.
An advantage of \$PINN from its domain decomposition feature is its parallelization capacity. This would possibly reduce the training costs and outperform a model like BPINN, which is computationally heavy. Parallel computation is not carried out in this paper, however, future work is focused on this implementation. 

\bibliographystyle{elsarticle-num}
\bibliography{biblio}



\end{document}